\documentclass[letterpaper]{article} 
\usepackage{aaai23}  
\usepackage{times}  
\usepackage{helvet}  
\usepackage{courier}  
\usepackage[hyphens]{url}  
\usepackage{graphicx} 
\usepackage{natbib}  
\usepackage{caption} 
\usepackage{algorithm}
\usepackage{algorithmic}

\usepackage{newfloat}
\usepackage{listings}
\DeclareCaptionStyle{ruled}{labelfont=normalfont,labelsep=colon,strut=off} 
\floatstyle{ruled}
\newfloat{listing}{tb}{lst}{}
\floatname{listing}{Listing}
\title{VASR: Visual Analogies of Situation Recognition}
\author {
    Yonatan Bitton,
    Ron Yosef, 
    Eli Strugo,
    Dafna Shahaf,
    Roy Schwartz,
    Gabriel Stanovsky
}
\affiliations {
    The Hebrew University of Jerusalem\\
    \{yonatan.bitton,ron.yosef,eli.strugo,dafna.shahaf,roy.schwartz1,gabriel.stanovsky\}@mail.huji.ac.il
}
\usepackage{bibentry}

\usepackage{booktabs}
\usepackage{multirow}
\usepackage{caption}
\usepackage{subcaption}
\newcommand\extrafootertext[1]{%
    \bgroup
    \renewcommand\thefootnote{\fnsymbol{footnote}}%
    \renewcommand\thempfootnote{\fnsymbol{mpfootnote}}%
    \footnotetext[0]{#1}%
    \egroup
}

\newcommand{\zeroshot}[0]{\emph{Zero-Shot Arithmetic}}
\newcommand{\trainedarithmetic}[0]{\emph{Supervised Arithmetic}}
\newcommand{\trainedconcat}[0]{\emph{Supervised Concat}}

\begin{document}

\maketitle

\begin{abstract}
A core process in human cognition is \emph{analogical mapping}: the ability to identify a similar relational structure between different situations.
We introduce a novel task, Visual Analogies of Situation Recognition, adapting the classical word-analogy task into the visual domain. Given a triplet of images, the task is to select an image candidate B' that completes the analogy (A to A' is like B to what?). Unlike previous work on visual analogy that focused on simple image transformations, we tackle complex analogies requiring understanding of scenes. 

We leverage situation recognition annotations and the CLIP model to generate a large set of 500k candidate analogies. Crowdsourced  annotations for a sample of the data indicate that humans agree with the dataset label $\sim$80\% of the time (chance level 25\%). Furthermore, we use human annotations to create a gold-standard dataset of 3,820 validated analogies.
Our experiments demonstrate that state-of-the-art models do well when distractors are chosen randomly ($\sim$86\%), but struggle with carefully chosen distractors ($\sim$53\%, compared to 90\% human accuracy). We hope our dataset will encourage the development of new analogy-making models. Website: \url{https://vasr-dataset.github.io/}
\end{abstract}

\section{Introduction}
The ability to draw analogies, flexibly mapping relations between superficially different domains, is fundamental to human intelligence, creativity and problem solving \cite{hofstadter2013surfaces,depeweg2018solving,goodman2014concepts,fauconnier1997mappings,gentner2001analogical,carey2011precis,spelke2007core}. This ability has also been suggested to be key to constructing more general and trustworthy AI systems \cite{mitchell2021abstraction,mccarthy2006proposal}. 
\begin{figure}[!tb]
\centering
\newcommand{\figlen}[0]{\columnwidth}
    \includegraphics[width=\figlen]{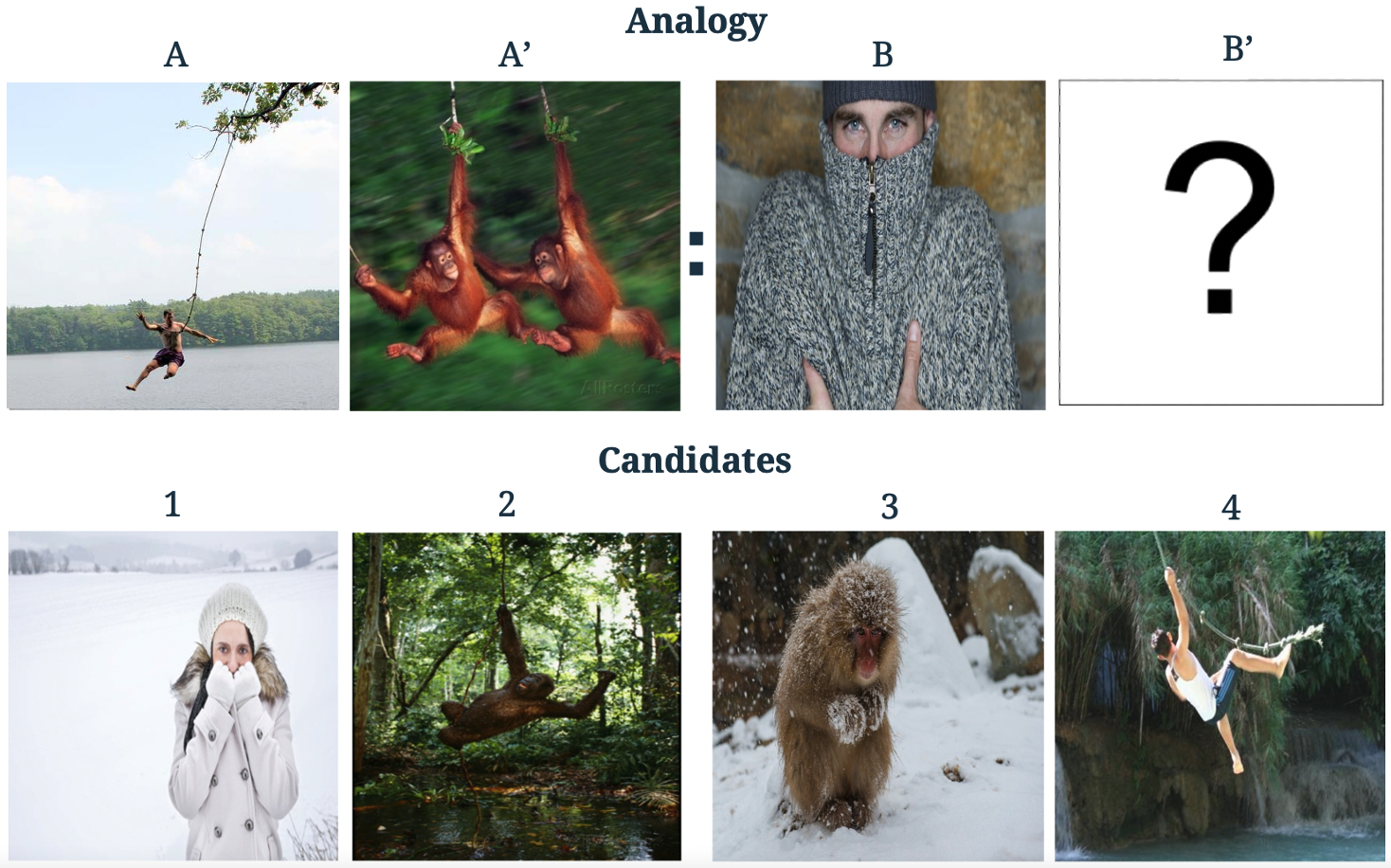}\\
    \caption{An example of visual analogy from the VASR dataset. The task is to select an image which best completes the analogy. The answer is found in the footnote.}
    \label{fig:fig1}
\end{figure}
An essential part of analogical thinking is the ability to 
look at different \emph{situations} and extract abstract patterns. For example, a famous analogy is between the solar system and the Rutherford-Bohr model of the
atom. Importantly, while the surface features are very different (atoms are much smaller than planets, different forces are involved, etc.), both phenomena share deep structural similarity (e.g., smaller objects revolving around a massive object, attracted by some force).

Most computational analogy works to date have focused on text~\cite{mikolov2013linguistic,allen2019analogies}, often studying SAT-type analogies (e.g., walk:legs :: chew:mouth). In works involving analogies between \emph{situations} \cite{falkeneheimer1986structure,evans1964program,winston1980learning,gentner1983structure}, both entities and relations need explicit structured representations, limiting their scalability. 
In the visual domain, works also focused on SAT-type questions~\cite{lovett2017modeling,lake2015human,depeweg2018solving}, synthetic images~\cite{lu2019seeing,reed2015deep} or images depicting static objects, where the analogies focus on object properties (color, size, etc.)~\cite{tewel2021zero,sadeghi2015visalogy}, rather than requiring understanding of a full scene.

\extrafootertext{Answer: $3$. Between A and A', \textit{man} changed to \textit{monkey}. Thus, from B to B', a \textit{man} feeling cold changes to a \textit{monkey} feeling cold.}

In this work we argue that images are a promising source of \textit{relational} analogies between situations, as they provide rich semantic information about the scenes depicted in them.
We take a step in that direction and introduce the Visual Analogies of Situation Recognition (VASR) dataset. Each instance in VASR is composed of three images (A, A', B) and $K=4$ candidates (see Figure~\ref{fig:fig1}). The task is to select the candidate B’ such that the relation between B and B' is most analogous to the relation between A and A'. To solve the analogy in Figure~\ref{fig:fig1}, one needs to understand the key difference between A and A' (the main entity is changed from \textit{man} to \textit{monkey}) and map it to B (``A \textit{man} feeling cold'' is changed to ``A \textit{monkey} feeling cold''). Importantly, VASR focuses on situation recognition that requires understanding the full scene, the different roles involved and how they relate to each other.

\begin{figure}[!tb]
\centering
\newcommand{\figlen}[0]{\columnwidth}
    \includegraphics[width=\figlen]{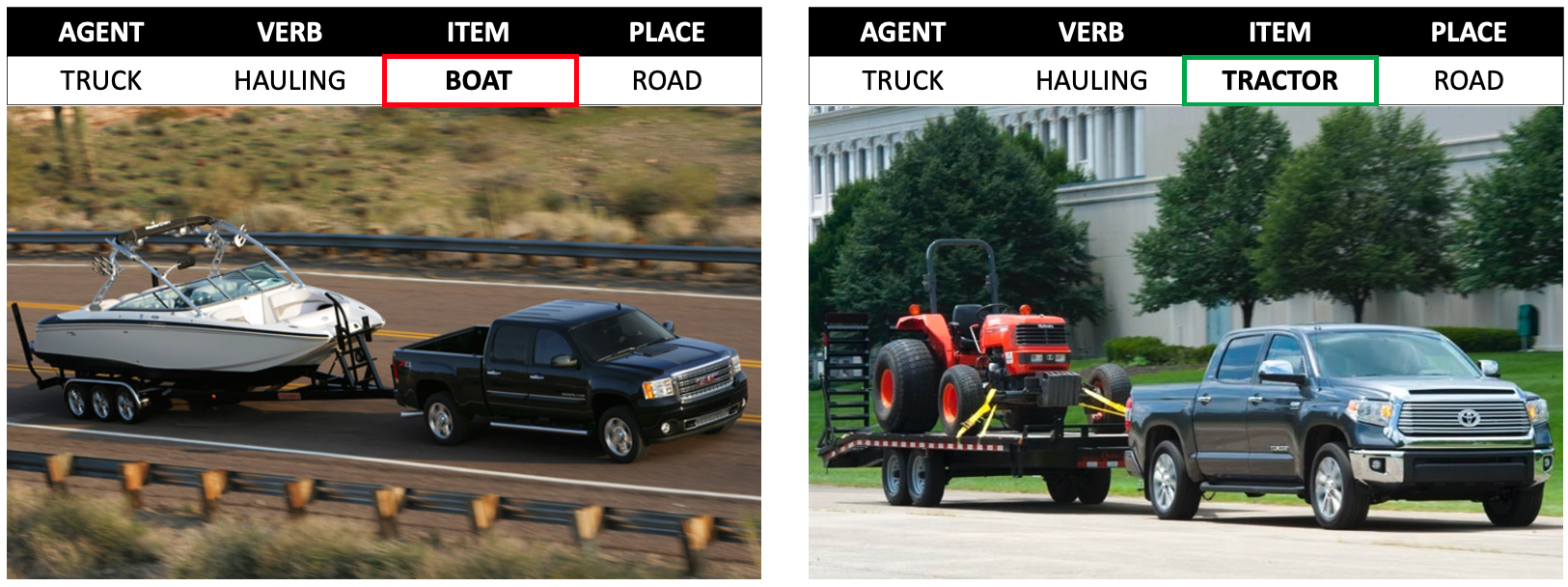}\\
    \caption{Two images and their situation recognition annotations from imSitu. In this example, both images share the same annotations except for the \emph{item} role (\emph{boat} $\rightarrow$ \emph{tractor}).}
    \label{fig:fig_AB}
\end{figure}

To create VASR, we develop an automatic method that leverages situation recognition annotations\footnote{Often referred to as visual semantic role labeling~\cite{gupta2015visual}.} to generate silver analogies of different kinds.\footnote{We use the term ``silver labels'' to refer to labels generated by an automatic process, which, unlike gold labels, are not validated by human annotators.} We start with the imSitu corpus~\cite{yatskar2016situation}, which annotates frame roles in images. For example, in the image on the left of Figure~\ref{fig:fig_AB}, the \emph{agent} is a \emph{truck}, the \emph{verb} is \emph{hauling}, and the \emph{item} (or \textit{theme}) is a \emph{boat}. We search for instances $A:A' :: B:B'$ where: (1) $A:A'$ are annotated similarly except for a single different role; (2) $B:B'$ exhibit the same delta in frame annotation. For example in Figure~\ref{fig:fig_AB}, the images are annotated the same except for \emph{item} that is changed from \emph{boat} to \emph{tractor}. The corresponding $B:B'$ images pairs should similarly have \emph{boat} as an \emph{item} role in $B$, and \emph{tractor} as an \emph{item} in $B'$, while all other roles are identical between them. We use several filters aiming to keep pairs of images that have a single main salient difference between them, and carefully choose the distractors to adjust the difficulty of the task.
This process produces over 500,000 instances, with diverse analogy types (activity, tool being used, etc.).

To create a gold standard and to evaluate the automatic generation of VASR, we crowd-source a portion of 4,170 analogies of the silver annotations using five annotators. On the test set, we find that annotators are very likely (93\%) to agree on the analogy answer, and reach high agreement with the auto-generated label (79\%). For human evaluation, we crowd-source additional annotations from new annotators who did not participate in the data generation part, evaluating a sample of 10\% of the gold-standard test set, finding that they solve it with high accuracy (90\%).

We evaluate various state-of-the-art computer vision models (ViT \cite{dosovitskiy2020image}, Swin Transformer \cite{liu2021swin}, DeiT \cite{touvron2021training} and ConvNeXt \cite{liu2022convnet}) in zero-shot settings using arithmetic formulations, following similar approaches in text and in vision~\cite{mikolov2013linguistic}. We find that they can solve analogies well when the distractors are chosen randomly (86\%), but all struggle with well-chosen difficult distractors, achieving only 53\% accuracy on VASR, far below human performance. 
Interestingly, we show that training baseline models on the large silver corpus is comparable with zero-shot performance and far below human performance, leaving room for future research.

Our main contributions are: (1) we present the VASR dataset as a resource for evaluating visual analogies of situation recognition; (2) we develop a method for automatically generating silver-label visual analogies from situation recognition annotations; (3) we show that current state-of-the-art models are able to solve analogies with random candidates, but struggle with more challenging distractors.

\section{Related Work}
The VASR dataset is built using annotations of situation recognition from imSitu, described below. In addition, we discuss two works most similar to ours, which tackle different aspects of analogy understanding in images.

\paragraph{Situation Recognition.} Situation recognition is the task of predicting the different semantic role labels~(SRL) in an image. For example in Figure~\ref{fig:fig1}, image A depicts a frame where the \emph{agent} is a \emph{person}, the \emph{verb} is \emph{swinging}, the \emph{item} is a \emph{rope}, and the \emph{place} is a \emph{river}.
The imSitu dataset \cite{yatskar2016situation}  presented the task along with annotated images gathered from Google image search, and a model for solving this task. Each annotation in imSitu comprises of \emph{frames}~\cite{fillmore2003background}, where each noun is linked to WordNet \cite{miller1995wordnet}, and objects are identified in image bounding boxes.\footnote{Follow-up work~\cite{pratt2020grounded} added bounding boxes to imSitu.}
We use these annotations to automatically generate our silver analogy dataset.

\paragraph{Analogies.} 
Analogies have been studied in multiple contexts. Broadly speaking, computational analogy methods can be divided into symbolic methods, probabilistic program induction, and neural approaches \cite{mitchell2021abstraction}.

In the context of analogies between \emph{images}, there have been several attempts to represent \emph{transformations} between pairs of images \cite{memisevic2010learning,reed2015deep,hertzmann2001image,forbus2011cogsketch}. The transformations studied were usually stylistic (texture transfers, artistic filters) or geometric (topological relations, relative position and size, 3D pose modifications). 

More recently, DCGAN \cite{radford2015unsupervised} has shown capabilities of executing vector arithmetic on images of faces, e.g. (man with glasses - man without glasses + woman without glasses $\approx$ woman with glasses). Another work, focusing on zero-shot captioning~\cite{tewel2021zero}, presented a model based on CLIP and GPT-2 \cite{radford2019language} for solving visual analogies, where the input consists of three images and the answer is textual. We evaluate their model in our experiments. 

Perhaps most similar to our work is VISALOGY~\cite{sadeghi2015visalogy}. In this work, the authors construct two image analogy datasets---a synthetic one (using 3D models of chairs that can be rotated) and a natural-image one, using Google image search followed by manual verification. However, even in the natural-image case, the analogies in VISALOGY are quite restricted; images mostly contain a single main object (e.g., a dog) and analogies based on attributes (e.g., color) or action (e.g., run). The VASR dataset contains analogies that are much more expressive, requiring understanding the full scene (see Figure~\ref{fig:single_object_vs_scene} in Appendix~\ref{sec:appendix}).
Importantly, the VISALOGY dataset is not publicly available, which makes VASR, to the best of our knowledge, the only publicly available benchmark for visual situational analogies with natural images.

\section{The VASR Dataset}
To build the VASR dataset, we leverage situation recognition annotations from imSitu. We start by finding likely image candidates based on the imSitu gold annotated frames~(\S\ref{sec:finding_analogies}). We then search for challenging answer distractors~(\S\ref{sec:distractors}). Following, we apply several filters~(\S\ref{sec:filtering}) in order to keep pairs of images with a single salient difference between them. We then select candidates for the gold test set~(\S\ref{sec:candidates_gold}), and crowdsource the annotation of a gold dataset (\S\ref{sec:mturk}). Finally, we provide the dataset statistics (\S\ref{sec:statistics}).

\subsection{Finding Analogous Situations in imSitu}
\label{sec:finding_analogies}

\begin{figure}[!tb]
\centering
    \includegraphics[width=0.8\columnwidth]{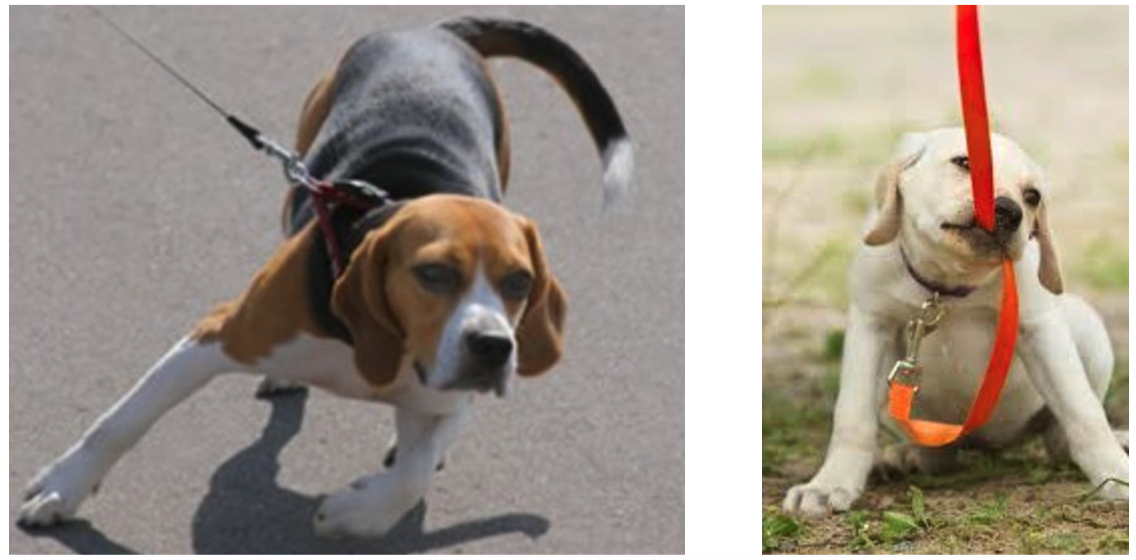}\\
    \caption{An image pair with \emph{multiple} salient visual differences (dog breed, activity, and more). We aim to filter these cases, keeping pairs with \emph{single} main salient difference.}
    \label{fig:fig_beagle_puppy}
\end{figure}

We start by considering the imSitu dataset containing situation recognition annotations of 125,000 images. We search for images $A:A'$ that are annotated the same, except for a single different role (e.g., the \emph{agent} role in Figure~\ref{fig:fig1} is changed from \emph{man} to \emph{monkey}). We extract image pairs that
have the same situation recognition annotation yet differ in one of the following roles: agent, verb, item, tool, vehicle and victim. This process yields $\sim$7 million image pairs. However, many of these pairs are not analogous because they do not have a \emph{single} salient visual difference between them (as exemplified in Figure~\ref{fig:fig_beagle_puppy}), due to partial annotation of the imSitu images. To overcome this, we apply several filters, described in Section~\ref{sec:filtering}, keeping $\sim$23\% of the pairs. Next, for each $A:A'$ pair we search for another pair of images, $B:B'$, which satisfy a single condition, namely that they exhibit the same difference in roles. Importantly, note that $B:B'$ can be very different from $A:A'$, as long as they adhere to this condition. 

\subsection{Choosing Difficult Distractors}
\label{sec:distractors}
\begin{figure}[!t]
\centering
\newcommand{\figlen}[0]{\columnwidth}
    \includegraphics[width=1\figlen]{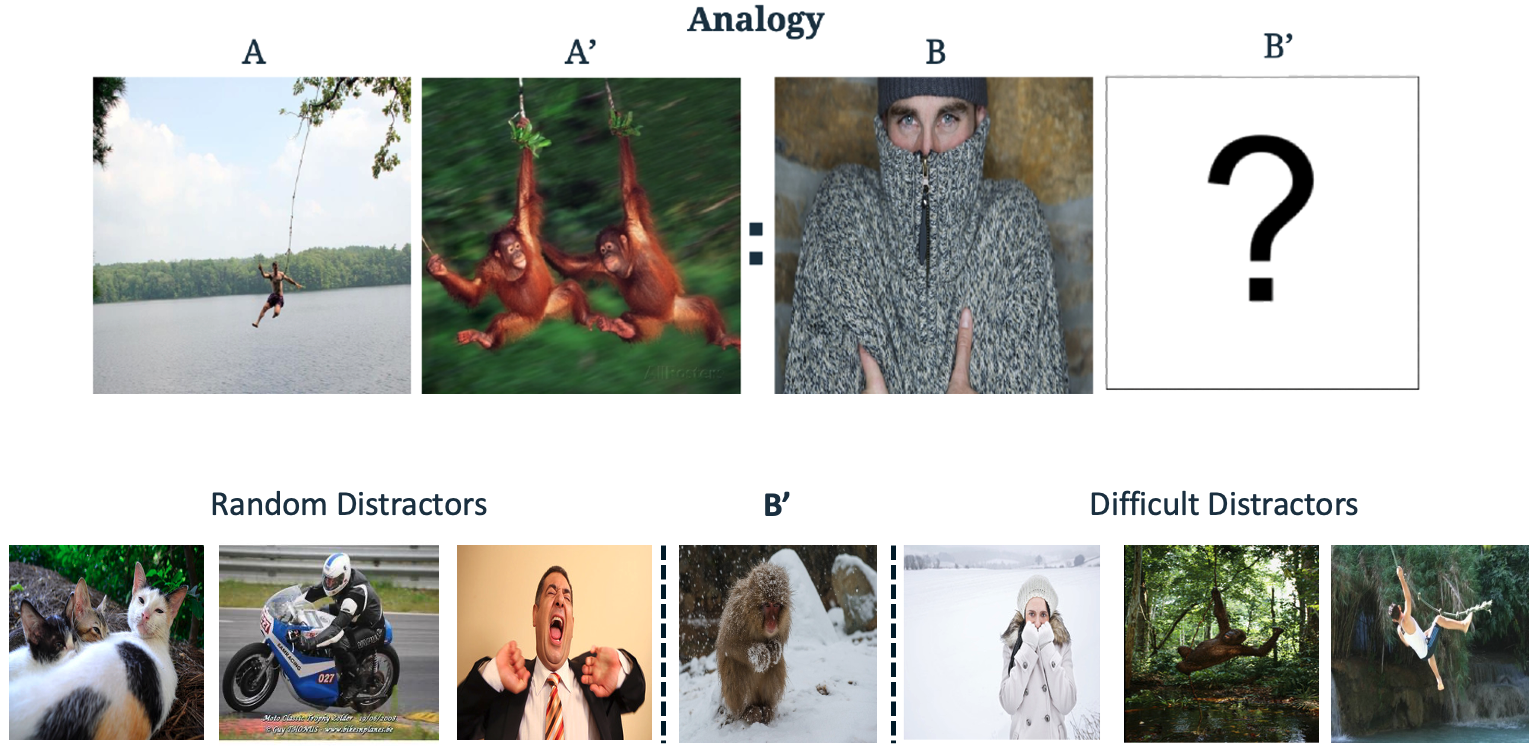}\\
    \caption{Compared to random distractors (on the left), VASR includes difficult distractors (on the right).}
    \label{fig:random_vs_difficult_distractors}
\end{figure}

Next, we describe how we compose VASR instances out of the analogy pairs collected in the previous section. The candidates are composed of the correct answer $B'$ and three other challenging distractors. Our experiments (\S\ref{sec:experiments}) demonstrate the value of our method for selecting difficult distractors compared to randomly selected distractors. Figure~\ref{fig:random_vs_difficult_distractors} illustrates this difference.

Specifically, we want distractors that would impede shortcuts as much as possible. Namely, the correct answer should involve two reasoning steps: (1) understanding the
key difference between $A:A'$ (the agent role \textit{man} changed to \textit{monkey} in Figure 1); (2) Map it to $B$. For the first reasoning step, we include distractors that are similar to $B$ but that do not have the same value in the changed role in $A'$ (candidates 1, 4 in Figure~\ref{fig:fig1} do \emph{not} depict a \emph{monkey}). For the second reasoning step, we include distractors with the changed role in $A'$ but in a different situation than B (candidate 2 in Figure~\ref{fig:fig1}, which does show a \emph{monkey}, but in a different situation). To provide such distractors, we search for images that are annotated similarly to $A'$ and $B$. For the similarity metric, we use an adaption of the Jaccard similarity metric between the images annotations. We calculate the number of joint values divided by the size of the union between the key sets of both images.\footnote{\url{https://en.wikipedia.org/wiki/Jaccard_index}. For example, for the two dictionaries \{ `a': 1, `b': 2 \}, \{ `a': 1, `c': 2 \}, the adapted Jaccard index is 1/3, because there is one joint value for the same key (`a': 1) and three keys in the union (`a',`b',`c')} We start by extracting multiple suitable distractors (40 in \emph{dev} and \emph{test}, 20 in \emph{train}). We later select the final 3 distractors using the filtering step described below (\S\ref{sec:filtering}).

\subsection{Filtering Ambiguous Image Pairs}
\label{sec:filtering}
We note that our automatic process is subject to several potential sources of error. One of them is the situation recognition annotations. The imSitu corpus was not created with analogies in mind, and as a result salient differences between the images are often omitted, and seemingly less important differences are highlighted. In this section, we attempt to ameliorate the issue and propose different filters to keep only pairs with one salient difference. We stress that there are many more filtering strategies possible, and exploring them is left for future work.

\subsubsection{Over-specified annotations}
We filter image pairs with overly-specific differences. For example, in Figure~\ref{fig:fig_beagle_puppy} the frames are annotated identically except for the \emph{agent} which is changed from \emph{beagle} to \emph{puppy}, while a human observer is likely to identify more salient differences (leash color, activity, and more). To mitigate such cases, we use a textual filter by leveraging imSitu's use of WordNet~\cite{miller1995wordnet} for nouns and FrameNet~\cite{fillmore2003background} for verbs. We identify the lowest common hypernyms for each annotated role (A \emph{beagle} is a type of a \emph{dog}, which is a type of a \emph{mammal}). Next, we only keep instances adhering to one of the following criteria: (1) both instances' corresponding roles are direct children to the same pre-defined WordNet concept class,\footnote{See full list of WordNet concepts in Appendix~\ref{sec:appendix}.} e.g., \emph{businessman} and \emph{businesswoman} are both direct children of \emph{businessperson}; (2) pairs of co-hyponyms, e.g., cat and dog are both animals, but a cat is not a dog and vice-versa; (3) the two instances belong to different clusters of animal, inanimate objects, or humans (e.g., \emph{bike} changed to \emph{cat} or \emph{car} changed to \emph{person}). This process removes $40\%$ of the original pairs. Filtered pairs are likely to be indistinguishable, for example: \emph{beagle} and \emph{puppy}, \emph{cat} and \emph{feline}, \emph{person} and \emph{worker}, and so on.

Another case of over-specific annotations is when a non visually salient object is being annotated. For example in Figure~\ref{fig:fig_non_visually_salient} in Appendix~\ref{sec:appendix} the annotated object is a small \emph{boomerang} that might be hard to identify. To mitigate such cases, we leverage bounding-boxes annotations from the SWiG dataset \cite{pratt2020grounded} and filter cases where the objects are hard to identify. Images with object size smaller than 2\% of the image size are filtered this way, filtering an additional 4\%.

\subsubsection{Under-specified annotations}
\begin{figure}[!t]
\centering
\begin{subfigure}{.49\textwidth}
  \centering
  \includegraphics[width=\linewidth]{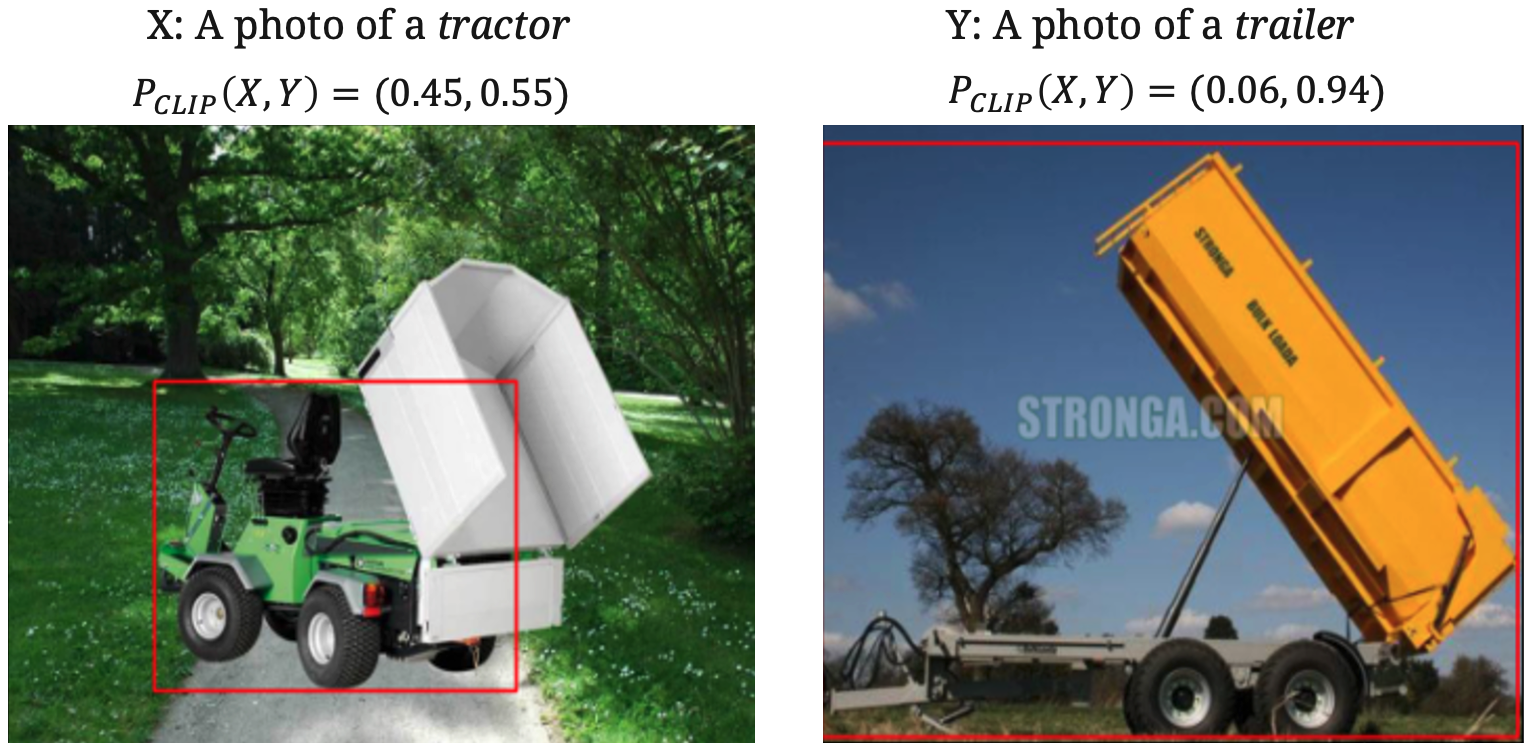}
  \caption{The left image bounding box is 55\% likely to be a photo of a \emph{trailer} rather than \emph{tractor}. Therefore we filter this case.}
\end{subfigure}%
\hspace{2mm}
\begin{subfigure}{.47\textwidth}
  \centering
  \includegraphics[width=\linewidth]{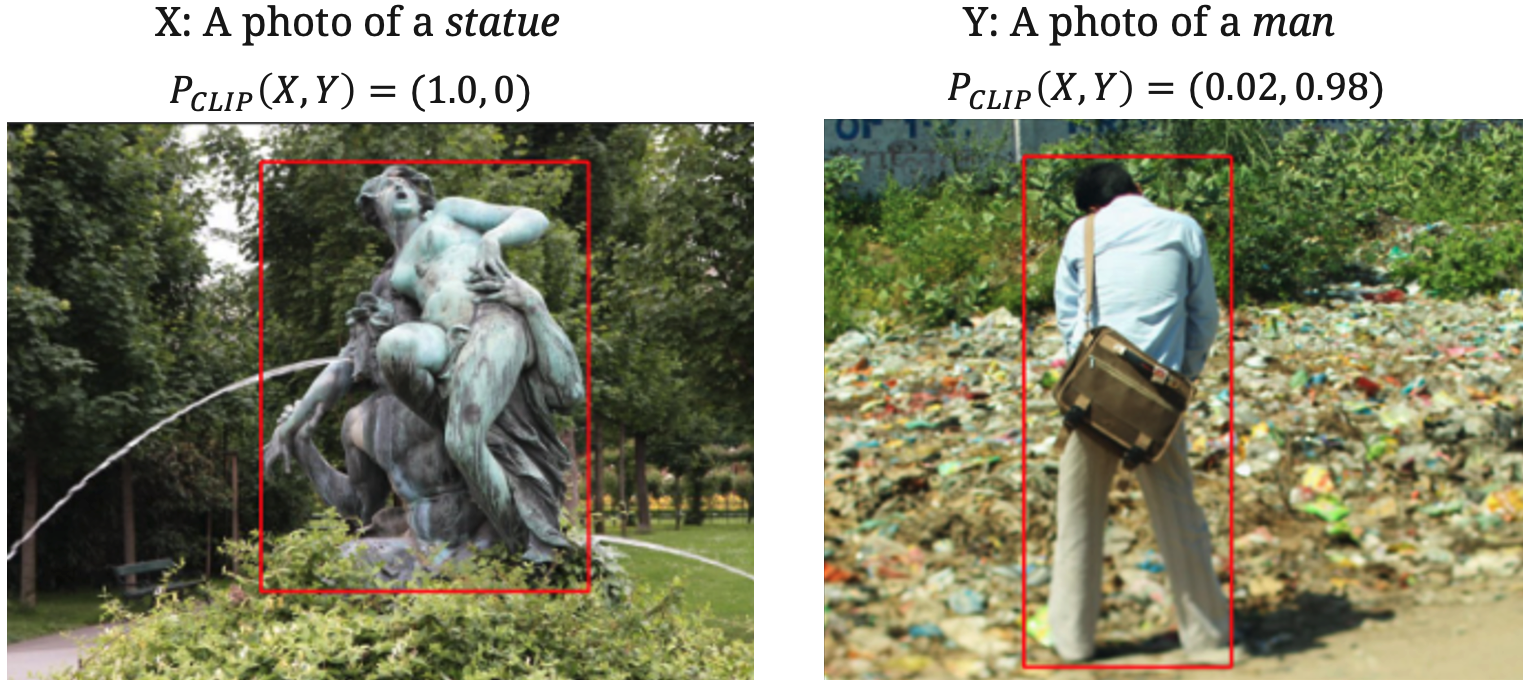}
  \caption{Both objects (\emph{statue}, \emph{man}) better describe their images bounding boxes (in 100\% and 98\%). Therefore we keep this instance.}
\end{subfigure}
\caption{Two examples for our CLIP based vision-and-language filtering. Given two images and annotated objects we compute the probabilities for each object to describe each image. We filter cases where an object can better describe the \emph{other} image rather than the image it annotates.}
\label{fig:clip_filter}
\end{figure}

The imSitu annotation is inherently bound to miss some information encoded in the image.
This can result in image pairs $A,A'$ that exhibit multiple salient differences, yet only a subset of them is annotated, leading to ambiguous analogies. For example in Figure~\ref{fig:clip_filter} top, the left image is described as a \emph{tractor}, and the right image described as a \emph{trailer}. However, the left image can be considered as a \emph{trailer} as well, and it is not clear what is the main difference between this images pair.
We aim to filter cases of such ambiguity, where an object can describe the \emph{other} image bounding box. For example, in Figure~\ref{fig:clip_filter}, the top example (a) is filtered by our method and the bottom example (b) is kept. 
Given two bounding boxes $X$, $Y$---each corresponding to different images---and two different annotated objects $X_{obj}$, $Y_{obj}$, we compute the CLIP \cite{radford2021learning} probabilities to describe each object bounding box using the prompt of ``A photo of a [OBJ]''. We denote  $$P_{X_{img}}(X_{obj},Y_{obj}) = (P(X_{img},X_{obj}), P(X_{img},Y_{obj}))$$ (and vice-versa for $Y$) and filter cases where it is not distinct. For example in the left image in Figure~\ref{fig:clip_filter}, $P_{X_{img}}(X_{obj},Y_{obj}) = (0.45, 0.55)$. The left image ($X$) is 55\% likely to be a photo of a \emph{trailer} ($Y$ annotation) rather than \emph{tractor} ($X$ annotation), therefore we filter this pair. We filter based on a threshold computed on a development set. We also execute a ``mesh filter'', where we combine all object labels of both images, measure the best object for each image, and filter cases where the best describing object for an image bounding box belongs to the other image. 

\begin{figure}[!t]
\begin{subfigure}{.48\textwidth}
  \includegraphics[width=\linewidth]{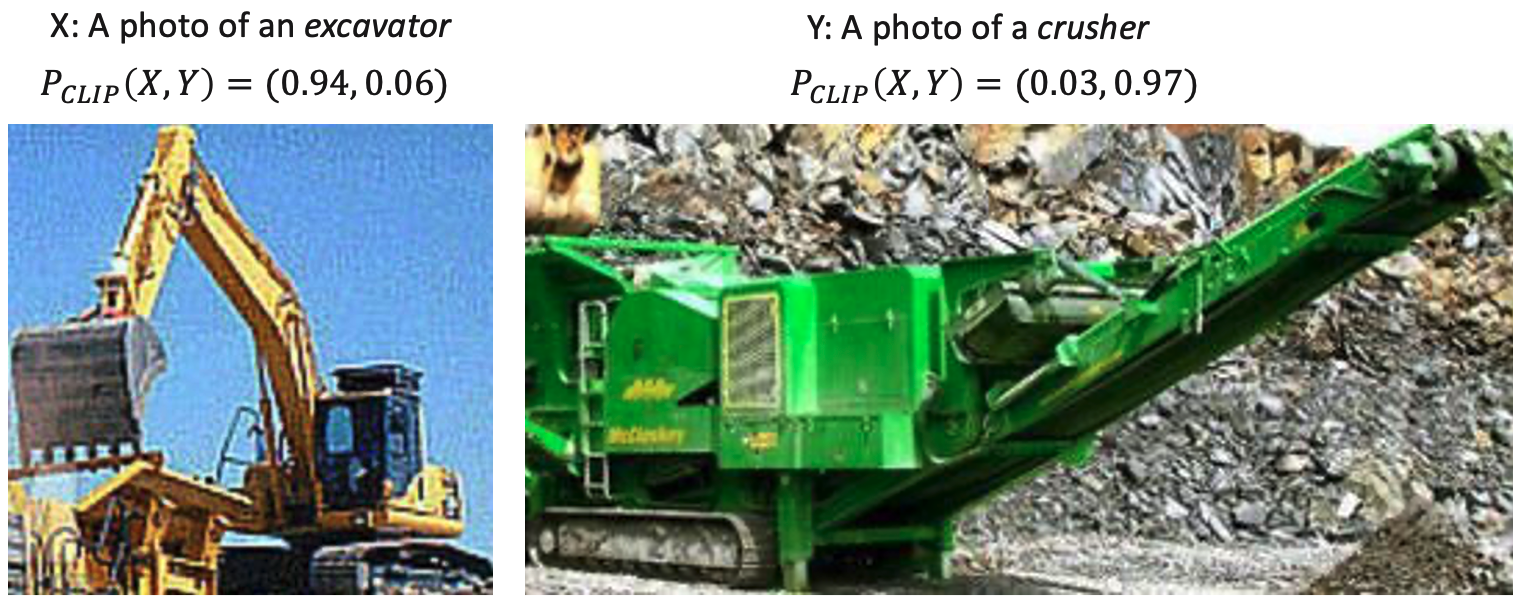}
  \caption{Based on the bounding box only, no ambiguity between the images and object classes.}
\end{subfigure}%
\hspace{2mm}
\begin{subfigure}{.48\textwidth}
  \includegraphics[width=\linewidth]{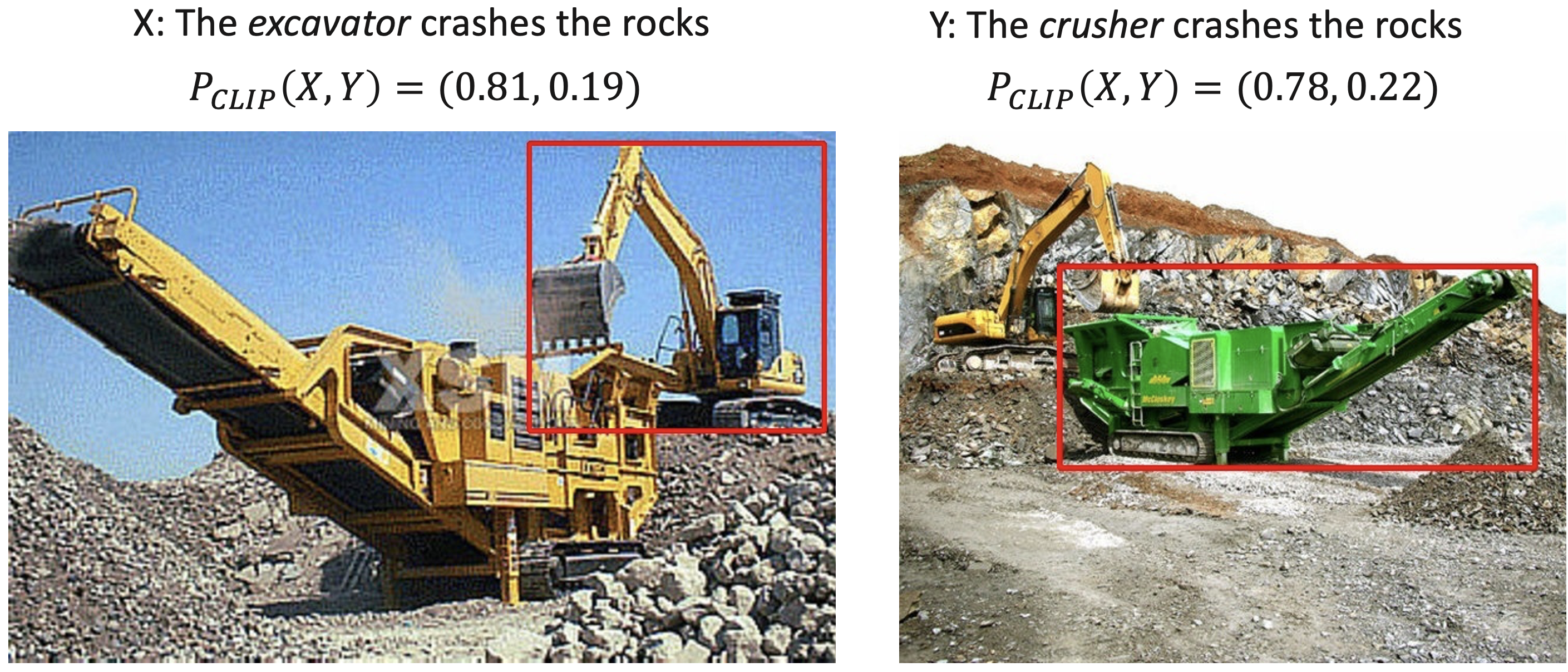}
  \caption{Based on the full image, the distinction between the images isn't that clear as in the bounding boxes case on the left.}
\end{subfigure}
\caption{CLIP-based filtering, bounding box vs. full image. The filter decision needs to consider both signals. Here the left figure is distinctive but the right is not, so we filter it out.}
\label{fig:clip_filter_full_img}
\end{figure}

Additionally to the objects and image bounding boxes, we  also take into consideration CLIP features extracted from the full image. Examples are presented in Figure~\ref{fig:clip_filter_full_img}. Instead of taking a template sentence of ``A photo of an [OBJ]'', we use a FrameNet template \cite{fillmore2003background} to receive a sentence describing the full image. For example the verb ``crashing'' (Figure~\ref{fig:clip_filter_full_img}) has the FrameNet template of: ``the AGENT crashes the ITEM\dots''. We substitute the annotated roles for the image, receiving a synthetic sentence describing the image. The CLIP probabilities are then used to filter indistinctive cases as in bounding-box filtering.

\subsection{Building the Test Set}
\label{sec:candidates_gold}
We aim to make the test set both challenging and substantially different from the training set in order to measure model generalizability. To do so, we select challenging test instances according to 3 metrics, defined below. In Section~\ref{sec:mturk}, we validate these instances via crowd-workers, finding them to be of good quality.
The metrics are: (1) an adapted Jaccard similarity metric to compute the difference in annotation between A, A'. We aim to select items with low Jaccard similarity to receive analogies that are \emph{distant} from each other; (2) calculate occurrences of each different key in the training set, in order to prefer rare items. For example $A:A'$ of \textit{girrafe : monkey} is preferred over \textit{man : monkey} if \textit{girrafe} appeared less than $man$ in the training set; (3) High annotation CLIP match: to avoid images with noisy annotations, we use the features computed in Section~\ref{sec:filtering} to calculate an ``Image SRL score'' using a weighted average of: (a) CLIP score of the caption to the image $P_{X_{img}}(X)$; (b) CLIP probability of the caption vs.~the caption from the other image pair. For example in the left image in Figure~\ref{fig:clip_filter} this score is 0.45. 
We sort our dataset according to these metrics, selecting 2,539 samples for the test set. We evaluate and annotate these candidates with human annotators (\S\ref{sec:mturk}).

\subsection{Human Annotation}
\label{sec:mturk}
We pay Amazon Mechanical Turk (AMT) crowdworkers to annotate the ground truth labels for a portion of VASR. We asked five annotators to solve 4,214 analogies.\footnote{To maintain high-quality work, we have a qualification task of 10 difficult analogies, requiring a grade of at least 90\% to enter the full annotation task. The workers received detailed instructions and examples from the project website.} Workers were asked to select the image that best solves the analogy, and received an estimated hourly pay of 12\$. Total payment to AMT was 1,440\$. Full details and examples of the AMT annotators screen are presented in Appendix~\ref{sec:appendix}, Section~\ref{sec:human_annotation_appendix}.

Table \ref{tab:mturk_results} shows some statistics of the annotation process. We observe several trends. First, in 93\% of the analogies there was an agreement of at least three annotators on the selected solution, compared to a probability of 41.4\% for a random agreement of at least three annotators on a any solution.\footnote{Binomial distribution analysis shows that the probability to get a random majority of at least 3 annotators out of 5 is 41.4\%.} Second, in 79\% of the instances the majority vote (of at least 3 annotators) agreed with the auto-generated dataset label. Moreover, given that the annotators reached a majority agreement, their choice is the same as the auto-generated label in 85\% of the cases. When considering annotators that annotated more than 10\% of the test set, the annotator with the highest agreement with the auto-generated label achieved 84\% agreement. Overall, these results indicate that the annotators are very likely to agree on a majority vote and with the silver label. 
The resulting dataset is composed of the 3,820 instances agreed upon with a majority vote of at least 3 annotators.

\begin{table}[tb!]
\center
\caption{AMT annotation results. The annotators are very likely to select the same candidate as the analogy answer, and with high agreement with the auto-generated label.}
\label{tab:mturk_results}
\resizebox{\columnwidth}{!}{\begin{tabular}{@{}lrrr@{}}
\toprule
                                                              & Test & Dev & Train \\ \midrule
\# samples fully annotated                                    & 2,539 & 178 & 1,492  \\
\% of samples with agreement of at least   three              & \textbf{93}   & 90  & 88    \\
\% of samples where majority vote agrees   with dataset label & \textbf{79}   & 75  & 75    \\ \bottomrule
\end{tabular}}
\end{table}
\subsection{Final Datasets and Statistics}
\label{sec:statistics}

The analogies generation process produces over 500,000 analogies using imSitu annotations. We used human annotators (\S\ref{sec:mturk}) to create gold-standard split, with 1,310, 160, 2,350 samples in the \emph{train}, \emph{dev}, \emph{test} (\S\ref{sec:candidates_gold}), respectively. Next, we create a silver \emph{train} of size 150,000 items and a silver \emph{dev} set of size 2,249 items. We sample the silver \emph{train} and \emph{dev} sets randomly, but we balance the proportions of different types of analogies similar to the \emph{test}.

VASR contains a total of 196,269 object transitions (e.g., \emph{book} changed to \emph{table}), of which 6,123 are distinct. It also contains 385,454 activity transitions (e.g., ``\emph{smiling}'' changed to ``\emph{jumping}''), of which 2,427 are distinct. Additional statistics are presented in Appendix~\ref{sec:appendix}, Section~\ref{sec:additional_stats}. To conclude, we have silver \emph{train} and \emph{dev} sets, and gold \emph{train, dev}, and \textit{test} sets. Full statistics are presented in Table~\ref{tab:fig_dataset_statistics}.

We encourage to focus on solving VASR with little or no training, since solving analogies requires mapping of existing knowledge to new, unseen situations~\cite{mitchell2021abstraction}. Evaluation of models should be performed on the (gold) \emph{test} set. To encourage development of models to solve VASR, an evaluation page is available on the website. The ground truth answers are kept hidden, predictions can be sent to our email and we will update the leaderboard. In a few-shot fine-tune setting, we suggest using the gold-standard \emph{train} and \emph{dev} splits, containing 1,470 analogies. For larger fine-tune, we suggest using the silver \emph{train} and \emph{dev} sets, with 152,249 analogies. We also publish the full generated data (over 500K analogies) to allow other custom splits. Next we turn to study state-of-the-art models' performance on VASR.

\begin{table}[!tb]
\caption{VASR statistics. Rows 1-2 describe the silver data, and rows 3-5 describe the gold-standard data.}
\label{tab:fig_dataset_statistics}
\centering
\resizebox{\columnwidth}{!}{\begin{tabular}{llrrrrrrr}
            \toprule
             & & Agent  & Verb   & Item   & Tool  & Vehicle & Victim & Total   \\ \midrule
\multirow{2}{*}{Silver} & Train & 82,984 & 38,331 & 20,836 & 6,360 & 1,343   & 146    & 150,000 \\
  & Dev   & 1,704  & 123    & 238    & 146   &         & 38     & 2,249   \\ \midrule
\multirow{3}{*}{Gold} & Train   & 558    & 116    & 376    & 170   & 90      &        & 1,310   \\
 & Dev     & 129    & 7      & 12     & 10    &         & 2      & 160     \\
 & Test    & 795    & 368    & 554    & 160   & 169     & 304    & \textbf{2,350}  
\\ \bottomrule
\end{tabular}}
\end{table}

\section{Experiments}
\label{sec:experiments}
We evaluate humans and state-of-the-art image recognition models in both zero-shot and supervised settings. We show that VASR is easy for humans (90\% accuracy) and challenging for models ($<$55\%). We provide a detailed analysis per analogy type, experiments with partial inputs (when only one or two images are available from the input), and experiments with increased numbers of distractors.


\subsection{Human Evaluation}
We sample 10\% of the test set, and ask annotators that did not work on previous VASR tasks to solve the analogies. Samples from the validation process are presented in Appendix~\ref{sec:appendix}, Section~\ref{sec:additional_examples}. Each analogy is evaluated by 10 annotators and the chosen answer is the majority of 6 annotators.\footnote{The probability to receive a random majority vote of at least six annotators out of 10 is 7.9\%.} We find that the human performance on the test set is 90\%. Additionally, in 93\% of the samples there was an agreement of at least six annotators. This high human performance indicates the high
quality of our end-to-end generation pipeline. 

\subsection{Zero-Shot Models} 
We compare four model baselines:
\begin{enumerate}
    \item \zeroshot{}: Inspired by Word2Vec~\cite{mikolov2013linguistic}, we extract visual features from pre-trained models for each image and represent the input in an \emph{arithmetic} structure by taking the embedding of $B+A'-A$. We compute its cosine similarity to each of the candidates and pick the most similar. We experiment with the following models: ViT \cite{dosovitskiy2020image}, Swin Transformer \cite{liu2021swin}, DeiT \cite{touvron2021training} and ConvNeXt \cite{liu2022convnet}.\footnote{The exact versions we took are the largest pretrained versions available in \emph{timm} library: ViT Large patch32-384, Swin Large patch4 window7-224, DeiT Base patch16 384, ConvNeXt Large.} Figure~\ref{fig:zero_shot_arithmetics} in Appendix~\ref{sec:appendix} illustrates this baseline.
    \item \emph{Zero-Shot Image-to-Text} \cite{tewel2021zero}  presented a model  for  solving visual analogy tests in zero-shot setting. Given an input of three images $A$,$A'$,$B$, this model uses an initial prompt (``An image of a \dots'') and generates the best caption for the image represented by the same \emph{arithmetic} representation we use: $B+A'-A$. We calculate the  CLIP score between each image candidate and the caption generated by the model, and select the candidate with the highest score.
    \item \emph{Distractors Elimination}: similar to a multi-choice quiz elimination, this strategy takes the three candidates that are most similar to the inputs $A, A', B$, eliminates them, and selects the last candidate as the final answer. We use the pre-trained ViT embeddings and compute cosine similarity in order to select the similar candidates. 
    \item \emph{Situation Recognition Automatic Prediction}: This strategy uses automatic situation recognition model prediction from SWiG \cite{pratt2020grounded}. It tries to find a difference between $A:A'$ in the situation recognition prediction and map it to $B$, in a reversed way to the VASR construction. For example in Figure~\ref{fig:fig1} it will select the correct answer \emph{if} both $A:A'$ and $B:B'$ are predicted with the same situation recognition prediction except $man$ changed to $monkey$.
\end{enumerate}

\subsection{Supervised Models} 
We also consider models fine-tuned on the silver data. We add a classifier on top of the pre-trained embeddings to select one of the 4 candidates. The first model baseline (denoted \trainedconcat{}) concatenates the input embeddings and learns to classify the answer ($A$,$A'$,$B$) $\rightarrow$ $B'$. The second model baseline (denoted \trainedarithmetic{}) has the same input representation as \zeroshot{}. To classify an image out of 4 candidates, we follow the design introduced in SWAG \cite{zellers2018swag},\footnote{\url{https://huggingface.co/transformers/v2.1.1/examples.html?#multiple-choice}} which was used by many similar works \cite{sun2019dream,huang2019cosmos,liang2019new,dzendzik2021english}. Basically, each of the image candidates is concatenated to the inputs features, followed by a linear network activation and a classifier that selects one of the options. We use the Adam \cite{kingma2014adam} optimizer, a learning rate of 0.001, batch size of 128, and train for 5 epochs. We take the model checkpoint with the best silver \emph{dev} performance out of the 5 epochs, and use it for evaluation. Figure~\ref{fig:supervised_model} in Appendix~\ref{sec:appendix} illustrates this model.

\subsection{Results and Model Analysis}
\begin{table}[!tb]
\begin{center}
\caption{VASR test set accuracy for several baselines in zero-shot and training. Bold indicates best result in section.
}
\label{tab:mega_table}
\resizebox{\columnwidth}{!}{\begin{tabular}{@{}llllll@{}}
\toprule
Section                                                                                      & \multicolumn{2}{l}{Experiment}                                                                           & {\begin{tabular}[c]{@{}l@{}}Random   \\ Distractors\end{tabular}} & {\begin{tabular}[c]{@{}l@{}}Difficult   \\ Distractors\end{tabular}} & Row \\ \midrule
\multirow{7}{*}{Zero-Shot}                                                                   & \multirow{4}{*}{\begin{tabular}[c]{@{}l@{}}Zero-Shot\\Arithmetic\end{tabular}}     & ViT                & \textbf{86}     & 50.3        & 1   \\  
                                                                                             &                                                                                     & Swin               & \textbf{86}     & \textbf{52.9}        & 2   \\  
                                                                                             &                                                                                     & DEiT               & 77.7   & 47.2        & 3   \\  
                                                                                             &                                                                                     & ConvNeXt           & 79     & 51.2        & 4   \\ \cmidrule(l){2-6} 
                                                                                             & \multicolumn{2}{l}{\begin{tabular}[c]{@{}l@{}}Zero-Shot   \\ Image-to-Text\end{tabular}}                 & 70     & 38.9        & 5   \\ \cmidrule(l){2-6} 
                                                                                             & \multicolumn{2}{l}{Distractors Elimination}                                                              & 0.9    & 23.4        & 6   \\ \cmidrule(l){2-6} 
                                                                                             & \multicolumn{2}{l}{\begin{tabular}[c]{@{}l@{}}Situation Recognition\\ Automatic Prediction\end{tabular}} & 31     & 24.6       & 7   \\ \midrule
\multirow{2}{*}{\multirow{2}{*}{\begin{tabular}[c]{@{}l@{}}Training on\\the Silver Data\end{tabular}}}                                                                   & \multicolumn{2}{l}{Concat}                                                                        & \textbf{84.1}  & \textbf{54.9}       & 8   \\ \cmidrule{2-6} 
                                                                                             & \multicolumn{2}{l}{Arithmetic}                                                                 & 83.7  & 47.4     & 9   \\ \midrule
\multirow{4}{*}{Partial Inputs}                                                              & \multirow{2}{*}{Zero-Shot}                                                          & A'                 & \textbf{84.4}   & 45.8        & 10  \\  
                                                                                             &                                                                                     & B                  & 77.6   & 24.7        & 11  \\ \cmidrule(l){2-6} 
                                                                                             & \multirow{2}{*}{Supervised}                                                          & Single image       & 82.1      & 44.8           & 12  \\  
                                                                                             &                                                                                     & Pair of images     & 83.8      & \textbf{46.3}           & 13  \\ \midrule
                                                                                             Humans                                                                                       &                                                                                     &                    &      &  \textbf{90}           & 14 
\\ \bottomrule
\end{tabular}}
\end{center}
\end{table}
Table~\ref{tab:mega_table} shows our \textit{test} accuracy results.
Rows 1-7 show the zero-shot results.
The \zeroshot{} models (R1-R4) achieve the highest results, with small variance between the models, reaching up to 86\% with random distractors and around 50\% on the difficult ones. The \emph{Zero-Shot Image-to-Text} (R5) achieves lower accuracies on both measures (70\% and 38.9\%, respectively). The other two models perform at chance level for difficult distractors.\footnote{\emph{Distractors Elimination} strategy is particularly bad with random distractors, as it eliminates the 3 images closest to the input, whereas the solution is often closer to the inputs than random distractors.} To conclude, models can solve analogies in zero-shot well when the distractors are random, but struggle with difficult distractors.

Results on training on the silver data are presented in rows 8-9. \trainedconcat{} representation performs better than the \trainedarithmetic{}. Interestingly, its performance (54.9\%, R8) is only 2\% higher than the best zero-shot baseline (\zeroshot{}, R2), and still far from human performance (R14). This small difference might be explained by the distribution shift between the training data and the test data (\S\ref{sec:candidates_gold}), which might make the trained models over-rely on specific features in the training set. To test this hypothesis, we consider the ViT model's \textit{supervised} performance on the  \textit{dev} set, which, unlike the test set, was not created to be different than the training set. We observe \textit{dev} performance levels similar to the \textit{test} set (56.7\% with the difficult and 86.6\% with random distractors), which hints 
that models might struggle to capture the information required to solve visual analogies from supervised data.

\paragraph{Analysis per Analogy Type.} We study whether humans and models behave differently for different types of analogies. We examine the \textit{test} performance of both humans and the ViT-based models \zeroshot{} and \trainedconcat{} per analogy type (Table~\ref{tab:res_per_analogy_types}). Humans solve VASR above 80\% in all analogy types, except for \emph{tool} (66\%). 
The average performance of both models on all categories is around 50\%, except for the \emph{Agent} category, which seems to benefit most from supervision. We propose several possible explanations: First, \emph{Agent} is the most frequent class. This does not seem to be the key reason for this result, as the performance of the second most frequent category, \emph{Item}, is far worse. Second, \emph{Agent} is the most visually salient class and the model learns to identify it. This also does not seem to be the reason, because we see that the bounding-box proportion (objects proportions are in the second row\footnote{For example the ``person that is feeling cold'' in Figure~\ref{fig:fig1} (image $B$) takes $>$90\% of the image size.}) of the \emph{Vehicle} class (55\%) are larger than the \emph{Agent} class (44\%), but the performance on it is far worse. Finally, solving \emph{Agent} analogies could be the most similar task to the pre-training data of the models we evaluate, which mostly include images with a single class, without complex scenes and other participants (e.g., images from ImageNet \cite{deng2009imagenet}). This hypothesis, if correct, further indicates the value of our dataset, which contains many non-Agent analogies, to challenge current state-of-the-art models. We also find that the \zeroshot{} and \trainedconcat{} predict the same answer only in 40\% of the time. An oracle that is correct if either model is correct reaches an accuracy of 76\%, suggesting that these models have learned to solve analogies differently.
\begin{table}[!t]
\caption{Results per analogy types of humans and models baselines. The class with the highest/lowest accuracy for each model is in bold. Data Percentage is the proportion of each class from the \emph{gold} test. Objects Proportion is the mean object size divided by full image size.}
\label{tab:res_per_analogy_types}
\centering
\resizebox{\columnwidth}{!}{\begin{tabular}{@{}llllllll@{}}
\toprule
                      & Agent & Item & Verb & Victim & Vehicle & Tool & Total \\ \midrule
Data Percentage (\%)       & \textbf{34}    & 24   & 16   & 13     & 7       & 7    & 100   \\ 
Objects Proportion (\%)       & \textbf{44}    & 27   &    & 42     & 55       & 18    &    \\ \midrule

Humans                & 95    &\textbf{98}   & 85   & 84     & 83      & \textbf{66}   & 89.9  \\
Arithmetic Zero-Shot  & 50    & \textbf{48}   & 49   & 48     & 56      & \textbf{58}   & 50.3  \\
Trained Concatenation & \textbf{69}    & 50   & 44   & 52     & 46      & \textbf{44}   & 54.9  \\
 \bottomrule
\end{tabular}}\end{table}

\paragraph{Partial Inputs.} Ideally, solving analogies should not be possible with partial inputs. We experiment with ViT pre-trained embeddings in two setups: (1) A \emph{Zero-Shot} baseline, where the selected answer is the candidate with the highest cosine similarity to the image embeddings of $A'$ or $B$. For example in Figure~\ref{fig:fig1}, $A'$ depicts a ``monkey swinging'' and $B$ depicts a ``person shivering''. The candidates most similar to these inputs are $1$ and $2$, and both are incorrect solutions; (2) A \emph{supervised} baseline, which is the same as \trainedconcat{}, but instead of using all three inputs, we use a single or a pair of images: $A, A', B, (A,B), (A,A'), (A',B)$.
Results are presented in Table~\ref{tab:mega_table}, R10-R13. In \emph{Zero-Shot}, the strategy of choosing an image that is similar to $A'$ (R10) reaches close to the full inputs performance with random distractors, but much lower with the difficult distractors. With the \emph{supervised} baseline, we show the best setup of a single image ($B$, in R12) and a pair of images ($(A,'B)$, R13). We observe a similar trend to the zero-shot setting, concluding that it is difficult to solve VASR using partial inputs.

\paragraph{Performance in the Presence of more Distractors}Since VASR is generated  automatically, we can add more distractors and measure models' performance. We take the \textit{test} set with the ground-truth answer provided by the annotators and change the number of distractors hyperparameter from 3 to 7, adding distractors to each of the random and difficult distractors splits, changing chance level from 25\% to 12.25\%. 
We repeat the zero-shot experiments and present the results in Table~\ref{tab:more_distractors}. The ViT performance on the difficult distractors drops from $50.3\%$ to $27.7\%$, while on the random distractors the decline is much more moderate, from $86\%$ to $78.7\%$. We observe a similar trend for the other models. 
The large drop in performance on the difficult distractors further indicates the importance of a careful selection of the distractors. 
\begin{table}[!tb]
\caption{With random candidates, the models manage to cope even though the task becomes twice as difficult. However, the performance drop is  larger with difficult distractors.}
\label{tab:more_distractors}
\resizebox{\columnwidth}{!}{\begin{tabular}{@{}lllllll@{}}
\toprule
         & \multicolumn{2}{l}{4 Candidates}                                                                                                        & \multicolumn{2}{l}{8 Candidates}                                                                                                        & \multicolumn{2}{l}{\% Drop}                                                                                                                 \\ \midrule
Models   & \begin{tabular}[c]{@{}l@{}}Random\\      Distractors\end{tabular} & \begin{tabular}[c]{@{}l@{}}Difficult\\      Distractors\end{tabular} & \begin{tabular}[c]{@{}l@{}}Random\\      Distractors\end{tabular} & \begin{tabular}[c]{@{}l@{}}Difficult\\      Distractors\end{tabular} & \begin{tabular}[c]{@{}l@{}}Random\\      Distractors\end{tabular} & \begin{tabular}[c]{@{}l@{}}Difficult\\      Distractors\end{tabular} \\ \midrule
ViT      & 86                                                                & 50.3                                                                 & 78.7                                                              & 27.7                                                                 & \textbf{8\%}                                                               & \textbf{45\%}                                                                 \\
Swin     & 86                                                                & 52.9                                                                 & 78.2                                                              & 30.7                                                                 & 9\%                                                               & 42\%                                                                 \\
DeiT     & 77.7                                                              & 47.2                                                                 & 69.3                                                              & 27.1                                                                 & 11\%                                                              & 43\%                                                                 \\
ConvNeXt & 79                                                                & 51.2                                                                 & 70.2                                                              & 29.1                                                                 & 11\%                                                              & 43\%                                                                 \\ \bottomrule
\end{tabular}}
\end{table}

\section{Conclusions}
We introduced the VASR dataset for visual analogies of situation recognition. 
We automatically created over 500K analogy candidates, showing their quality via high inter-annotator agreement and their efficacy for training.
Importantly, VASR test labels are human-annotated with high agreement. We showed that state-of-the-art models can solve our analogies with random distractors, but struggle with harder ones.

\section*{Acknowledgements}
We would like to thank Timo Schick, Yanai Elazar, Leshem Choshen, Moran Mizrahi and Oren Sultan for their valuable feedback. This work was supported in part by the Center for Interdisciplinary Data Science Research at the Hebrew University of Jerusalem, and a research grant 2336 from the Israeli Ministry of Science and Technology. It was also supported by the European Research Council (ERC) under the European Union’s Horizon 2020 research and innovation programme (grant no. 852686, SIAM, Shahaf).

\bibliography{aaai23}

\section{Appendix}
\label{sec:appendix}

\subsection{License and Privacy}
All images we use are taken from the SWiG dataset \url{https://github.com/allenai/swig} licensed under the MIT license. The VASR dataset is thus also licensed under the MIT license. We do not collect or publish players personal information

\subsection{Reproducibility Checklist}

\subsubsection{Checklist}
\begin{enumerate}
    \item This paper Includes a conceptual outline and/or pseudocode description of AI methods introduced: \textbf{yes} 
    \item Clearly delineates statements that are opinions, hypothesis, and speculation from objective facts and results: \textbf{yes}
    \item Provides well marked pedagogical references for less-familiare readers to gain background necessary to replicate the paper: \textbf{yes}
    \item Does this paper make theoretical contributions? \textbf{no}
    \item Does this paper rely on one or more datasets? \textbf{yes}
    \item A motivation is given for why the experiments are conducted on the selected datasets: \textbf{yes}
    \item All novel datasets introduced in this paper are included in a data appendix: \textbf{yes}
    \item All novel datasets introduced in this paper will be made publicly available upon publication of the paper with a license that allows free usage for research purposes: \textbf{yes}
    \item All datasets drawn from the existing literature (potentially including authors’ own previously published work) are accompanied by appropriate citations: \textbf{yes}
    \item All datasets drawn from the existing literature (potentially including authors’ own previously published work) are publicly available: \textbf{yes}
    \item All datasets that are not publicly available are described in detail, with explanation why publicly available alternatives are not scientifically satisficing: \textbf{Not applicable}
    \item Does this paper include computational experiments? \textbf{yes}
    \item Any code required for pre-processing data is included in the appendix.
    \textbf{yes}
    \item All source code required for conducting and analyzing the experiments is included in a code appendix.
    \textbf{yes}
    \item All source code required for conducting and analyzing the experiments will be made publicly available upon publication of the paper with a license that allows free usage for research purposes.
    \textbf{yes}
    \item All source code implementing new methods have comments detailing the implementation, with references to the paper where each step comes from.
    \textbf{yes}
    \item If an algorithm depends on randomness, then the method used for setting seeds is described in a way sufficient to allow replication of results.
    \textbf{yes}
    \item This paper specifies the computing infrastructure used for running experiments (hardware and software), including GPU/CPU models; amount of memory; operating system; names and versions of relevant software libraries and frameworks.
    \textbf{yes}
    \item This paper formally describes evaluation metrics used and explains the motivation for choosing these metrics.
    \textbf{yes}
    \item This paper states the number of algorithm runs used to compute each reported result.
    \textbf{yes}
    \item Analysis of experiments goes beyond single-dimensional summaries of performance (e.g., average; median) to include measures of variation, confidence, or other distributional information.
    \textbf{yes}
    \item This paper lists all final (hyper-)parameters used for each model/algorithm in the paper’s experiments.
    \textbf{yes}

\end{enumerate}

\subsubsection{Models}
Models are described in Section~4.2. Zero-shot models run in less than two hours, and trained models in less than 24 hours, on a single Tesla K80 GPU. 
Trained models hyper-parameters are provided in Section~4.3. Full implementation is provided in the attached code. 

\subsubsection{Statistics}
Dataset generation method is described in Section~3. Statistics are provided in Section~3.6. A link to a downloadable version of the dataset is available in the code (install.sh file). Complete description of the annotation process is provided in Section~3.5. 

\subsubsection{Code}
Full implementation, dependencies, training code, evaluation code, pre-trained models, README files and commands to reproduce the paper results are provided in the attach code. 

\subsection{Additional VASR Examples}
\label{sec:additional_examples}

\begin{figure}[!h]
\centering
\newcommand{\figlen}[0]{\columnwidth}
    \includegraphics[width=0.95\figlen]{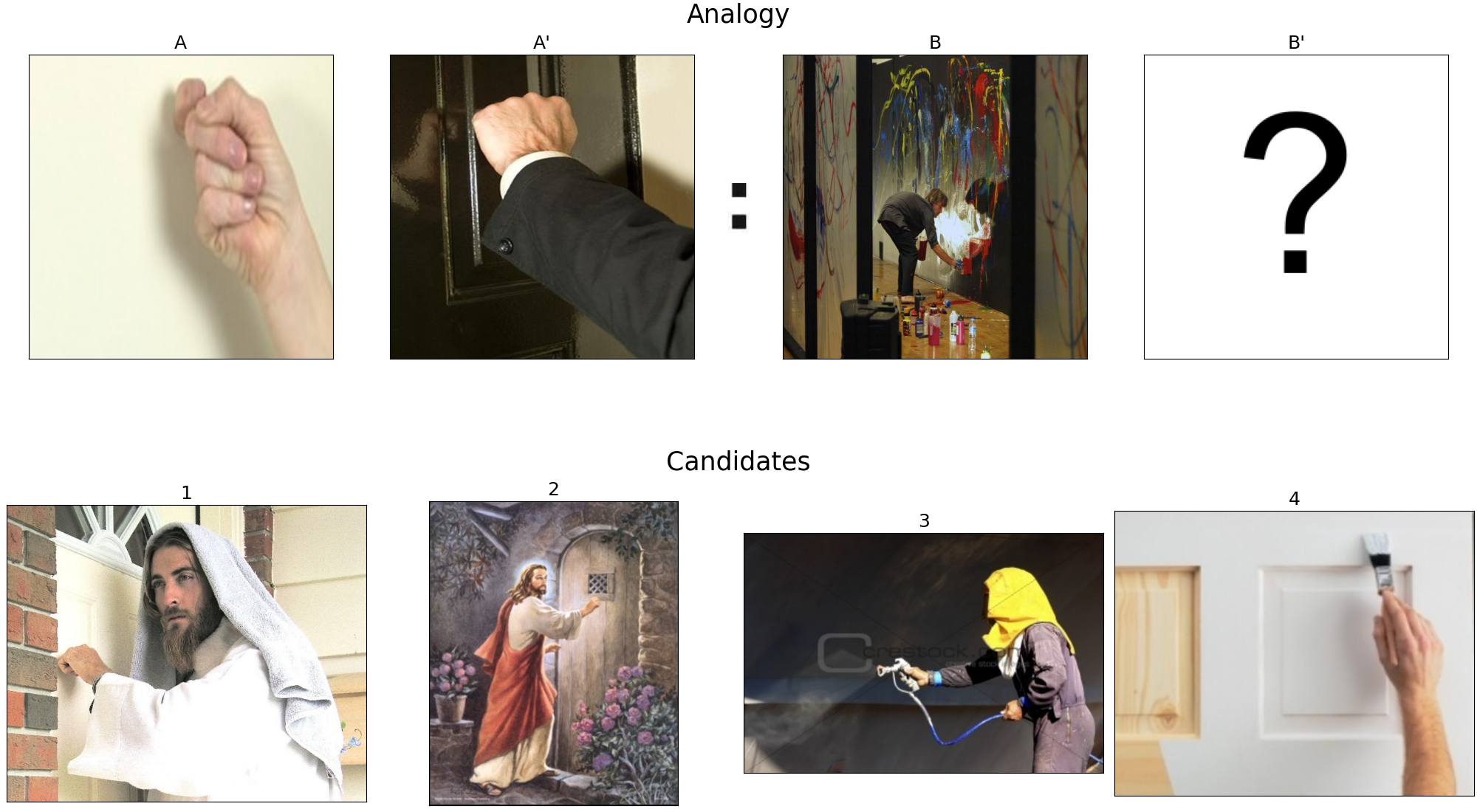}\\
    \caption{Answer - 4 (\emph{wall} changed to \emph{door})}
    \label{fig:e2}
\end{figure}

\begin{figure}[!h]
\centering
\newcommand{\figlen}[0]{\columnwidth}
    \includegraphics[width=0.95\figlen]{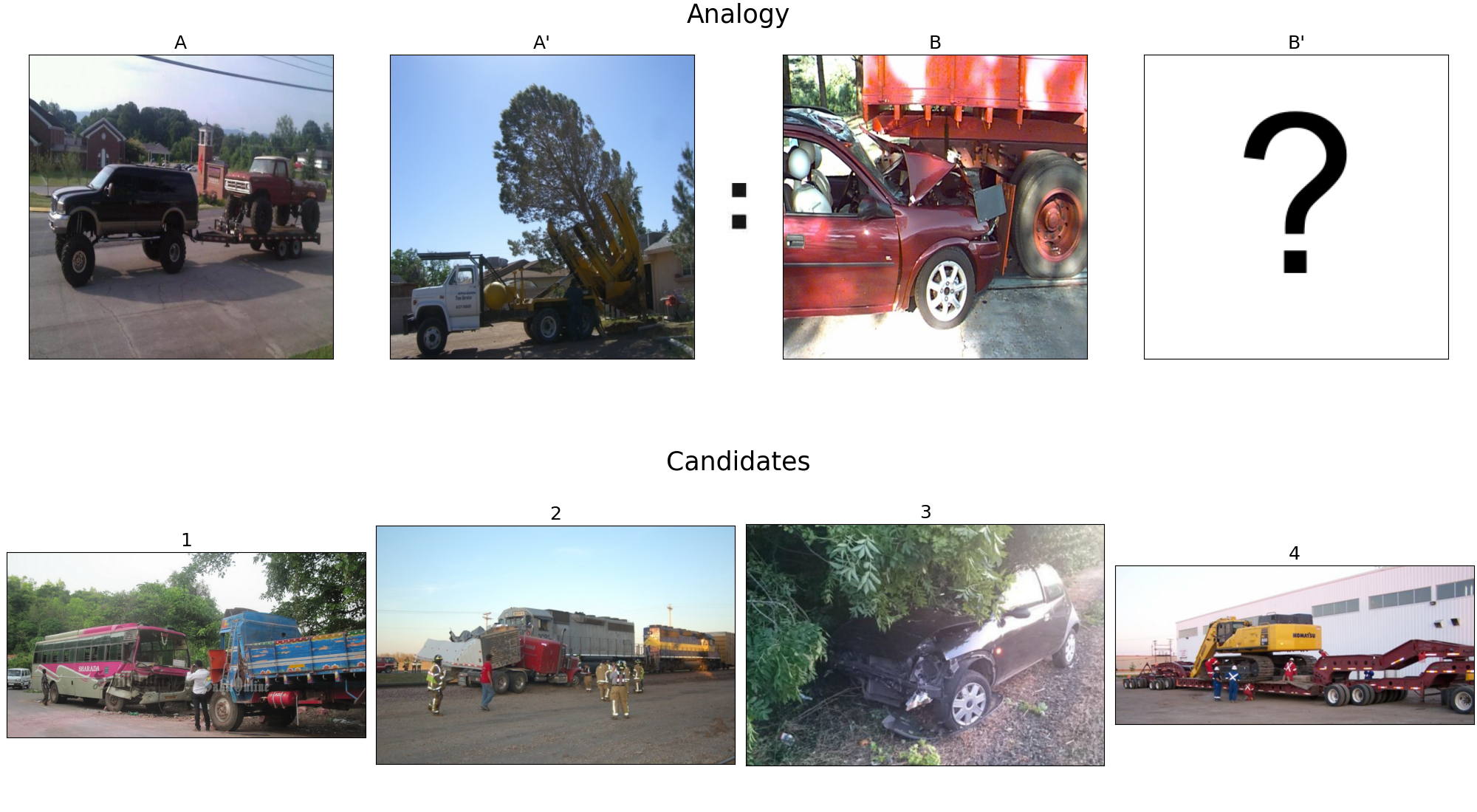}\\
    \caption{Answer - 3 (\emph{truck} changed to \emph{tree})}
    \label{fig:e3}
\end{figure}

\begin{figure}[!h]
\centering
\newcommand{\figlen}[0]{\columnwidth}
    \includegraphics[width=0.95\figlen]{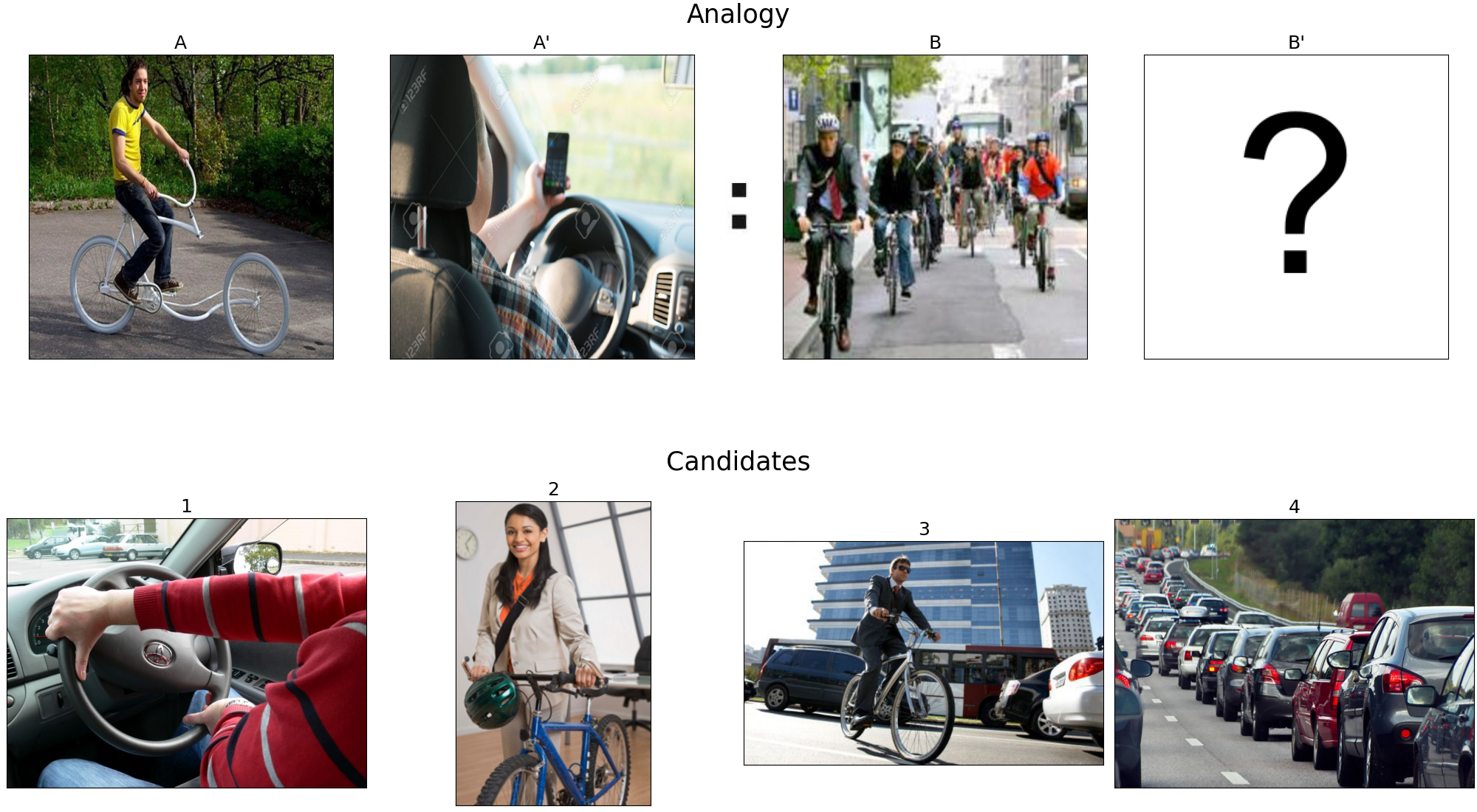}\\
    \caption{Answer - 4 (\emph{bicycle} changed to \emph{car})}
    \label{fig:e4}
\end{figure}

\begin{figure}[!h]
\centering
\newcommand{\figlen}[0]{\columnwidth}
    \includegraphics[width=0.95\figlen]{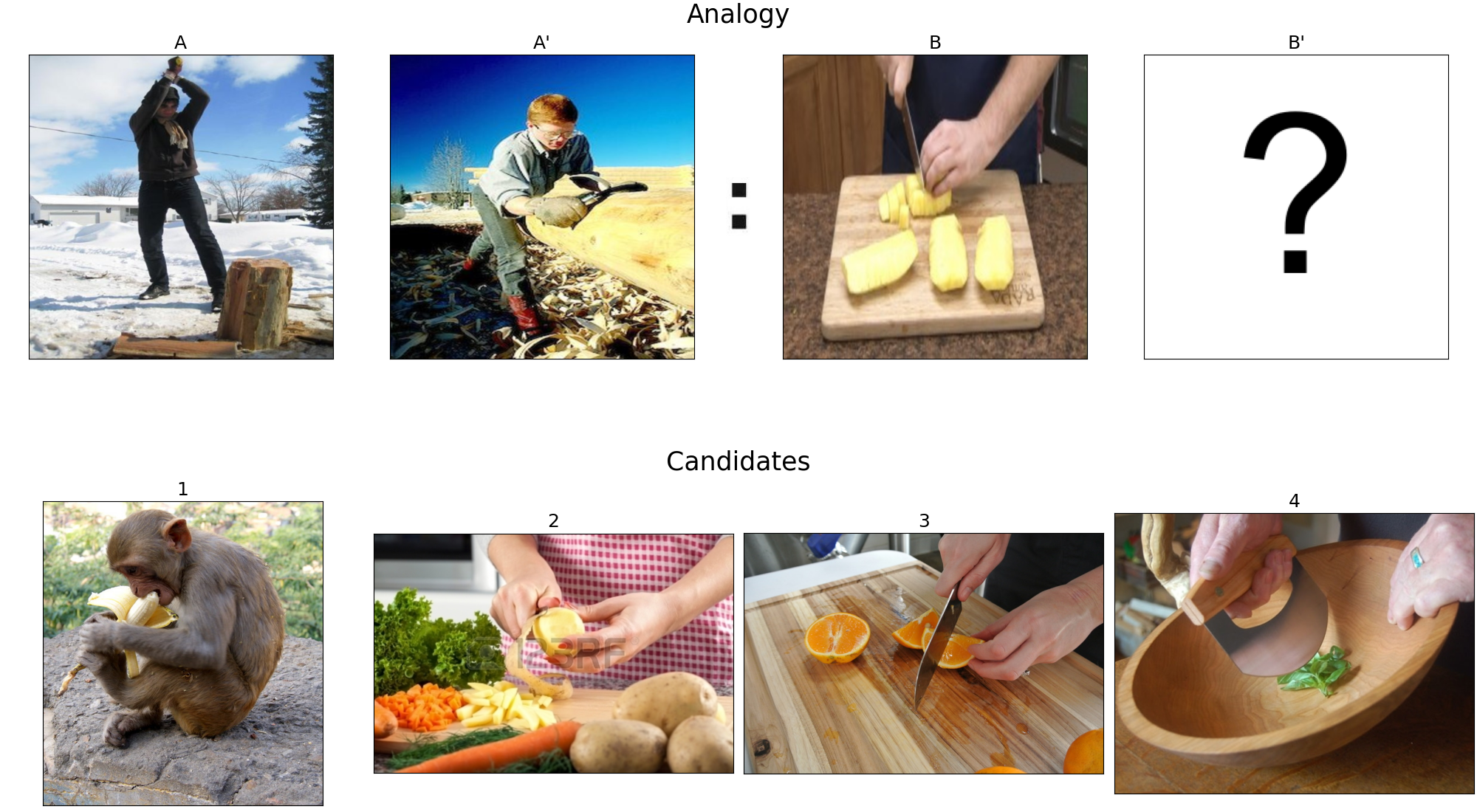}\\
    \caption{Answer - 2 (\emph{cut} changed to \emph{peel})}
    \label{fig:e6}
\end{figure}

\begin{figure}[!h]
\centering
\newcommand{\figlen}[0]{\columnwidth}
    \includegraphics[width=0.95\figlen]{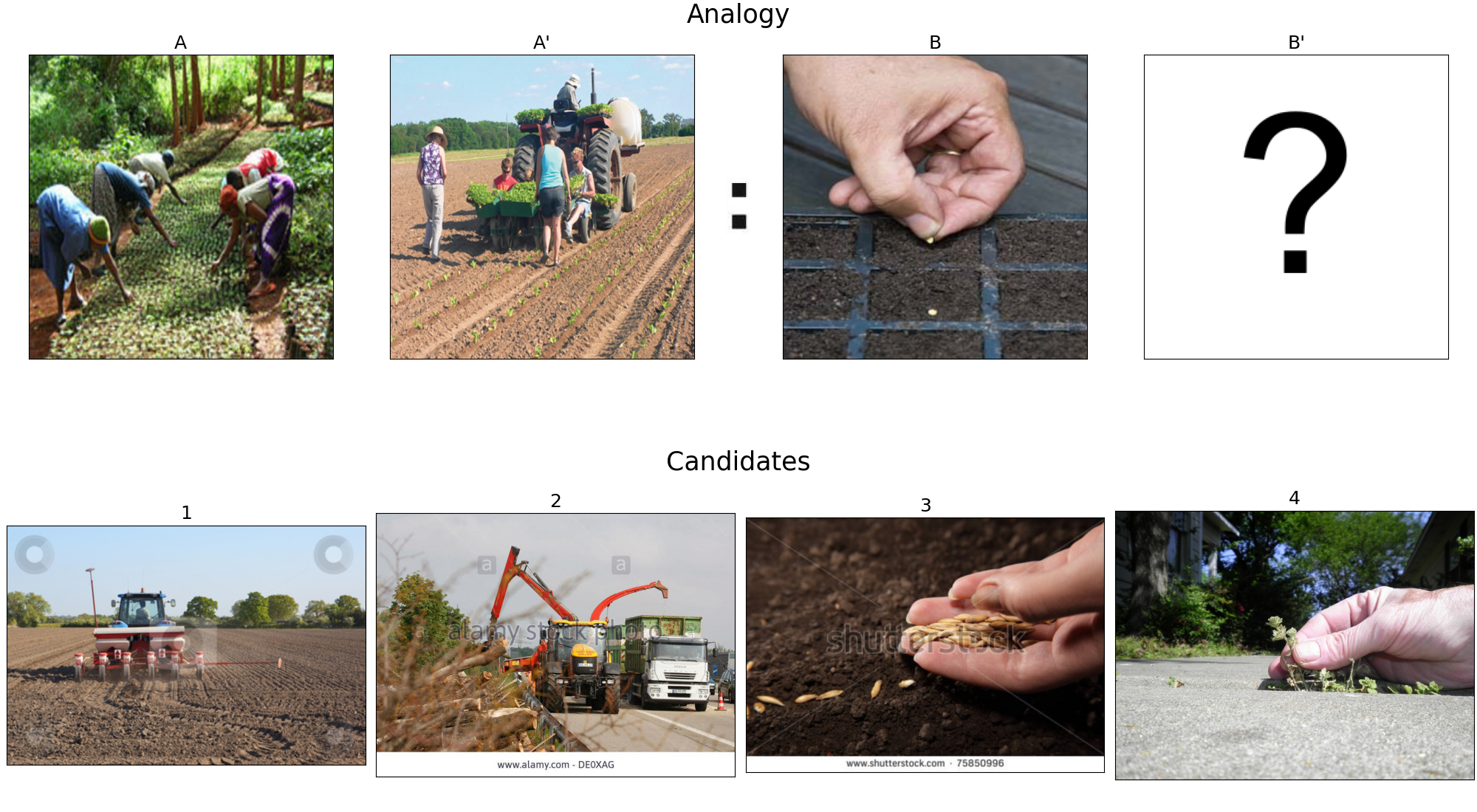}\\
    \caption{Answer - 1 (\emph{hand} changed to \emph{tractor})}
    \label{fig:e5}
\end{figure}

\begin{figure}[!h]
\centering
\newcommand{\figlen}[0]{\columnwidth}
    \includegraphics[width=0.95\figlen]{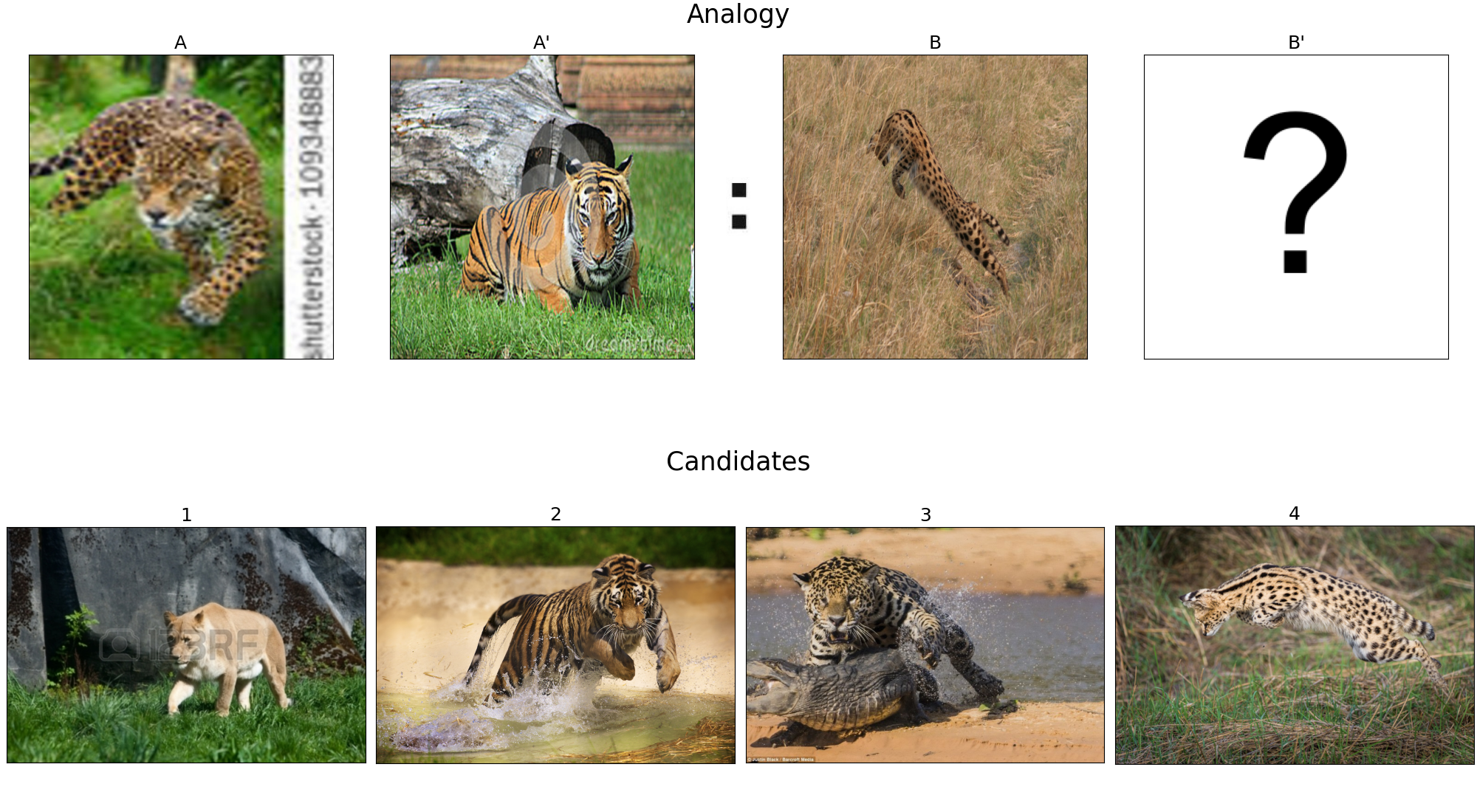}\\
    \caption{Answer - 2 (\emph{leopard} changed to \emph{tiger})}
    \label{fig:e1}
\end{figure}

\subsection{Human Annotation}
\label{sec:human_annotation_appendix}
\begin{figure}[!tb]
\centering
\newcommand{\figlen}[0]{\columnwidth}
    \includegraphics[width=\figlen]{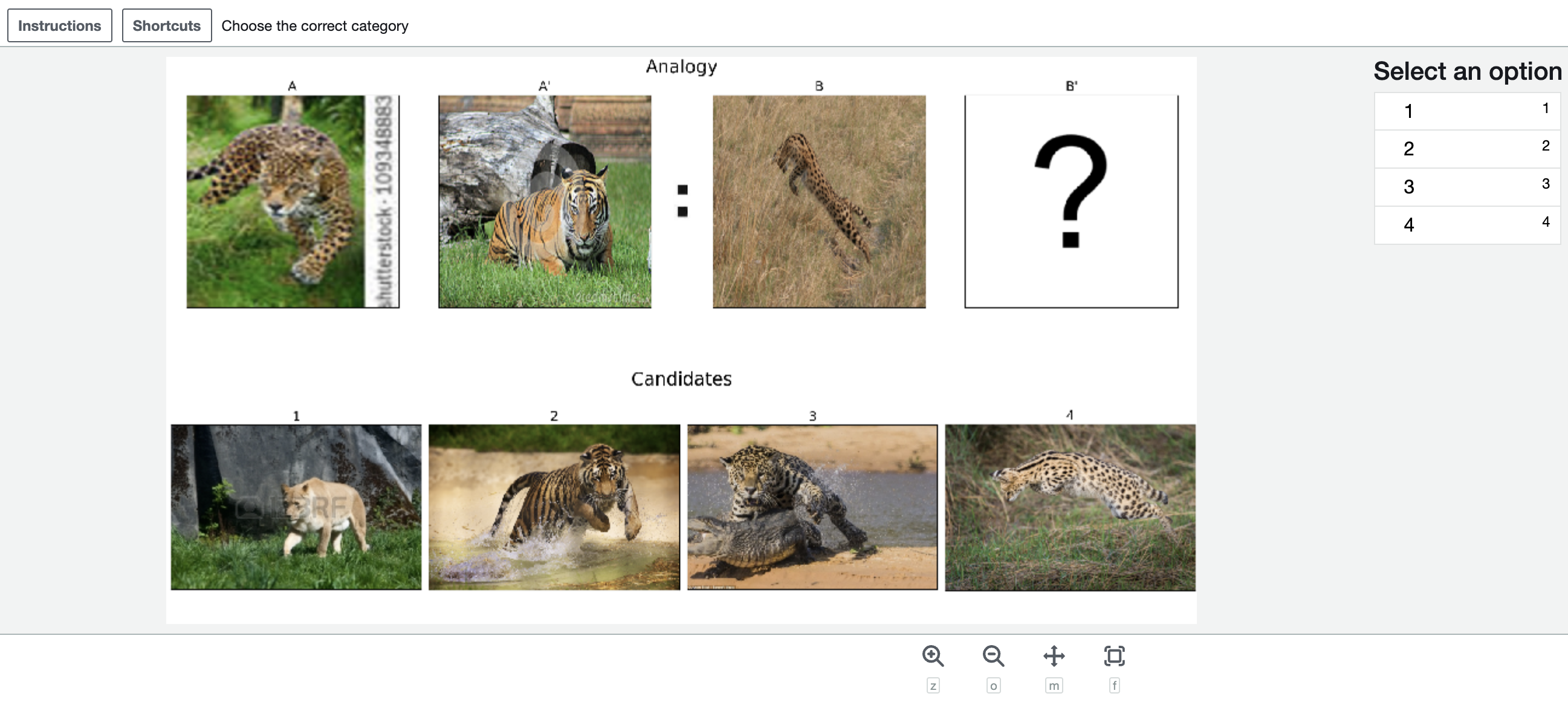}\\
    \caption{A screenshot from the annotator screen in Amazon Mechanical Turk.}
    \label{fig:mturk_example}
\end{figure}

Figure~\ref{fig:mturk_example} shows an example of the Mechanical Turk user-interface. The basic requirements for our annotation task is percentage of approved assignments above 98\%, more than 5,000 approved HITs. To be a VASR annotator, we required additional qualification tests: We selected 10 challenging examples from VASR as qualification test. To be qualified we accepted annotators that received a mean accuracy score over 90\%. The players received instructions (Section \ref{sec:instructions}) and could do an interactive practice in the project website.

\subsection{Annotators Instructions} 
\label{sec:instructions}
These are the instructions given to the annotators, accompanied by examples, and option to do an interactive practice in the project website:
``In the following you are expected to solve an analogy problem.
You will be shown three pictures: A, A', B.
There is some change going from picture A to picture A'
For example, A is a dog yawning and A' is a baby yawning - the change is dog → baby.

You need to choose an option out of 4 images.
Choose the image that best solves the analogy
A is to A' as B is to?

We recommend solving the analogies in computer, not mobile phone, as you'll need to see the images in large screen to succeed.

In addition, while you are in the HIT interface (after the qualification), we suggest to zoom-out (using Ctrl key and press the - [minus] key) in order to see the image in better resolution.

To enter the full task, there will be a qualification test which requires a score of 100

For additional (interactive!) examples, you may refer to the project website vasr-dataset.github.io.
Specifically, in the Explore Page you can learn on the different anlogies in the dataset, and in the Test Page you can test yourself on 5 analogies, receiving a score.

To solve it, understand what is the key difference between A and A', and map it to B.

It's possible to have several differences between A and A'. Search for the difference that allows you to choose a candidate that solves the analogy.

The difference between the images is one of the roles in the image:
(1) who is the agent in the image (man, horse, car, motorcycle, etc);
(2) the verb or the activity the agent is doing (e.g.,: a man nailing a nail);
(3) the tool the agent is using (e.g., a man nailing a nail with a hammer);
(4) the item that is effected by the agent (e.g., a man nailing a nail).''

\subsection{Additional Statistics} 
\label{sec:additional_stats}
\begin{table}[!h]
\begin{tabular}{@{}lllll@{}}
\toprule
        & \multicolumn{2}{l}{Silver} & \multicolumn{2}{l}{Gold} \\ \midrule
        & \# Total    & \# Unique    & \# Total   & \# Unique   \\
Objects & 113,585     & 3,490        & 3,329      & 1,315       \\
Verbs   & 38,664      & 1,989        & 491        & 160         \\ \bottomrule
\end{tabular}
\label{tab:more_stats}
\caption{Analogies Transitions Statistics. For example in Figure \ref{fig:fig1}, \emph{man} changed to \emph{monkey} is counted as a single object transition, and Figure \ref{fig:e6}, \emph{cut} changed to \emph{peel} is counted as a single verb transition.}
\end{table}

\begin{figure}[!h]
\centering
\newcommand{\figlen}[0]{\columnwidth}
    \includegraphics[width=0.95\figlen]{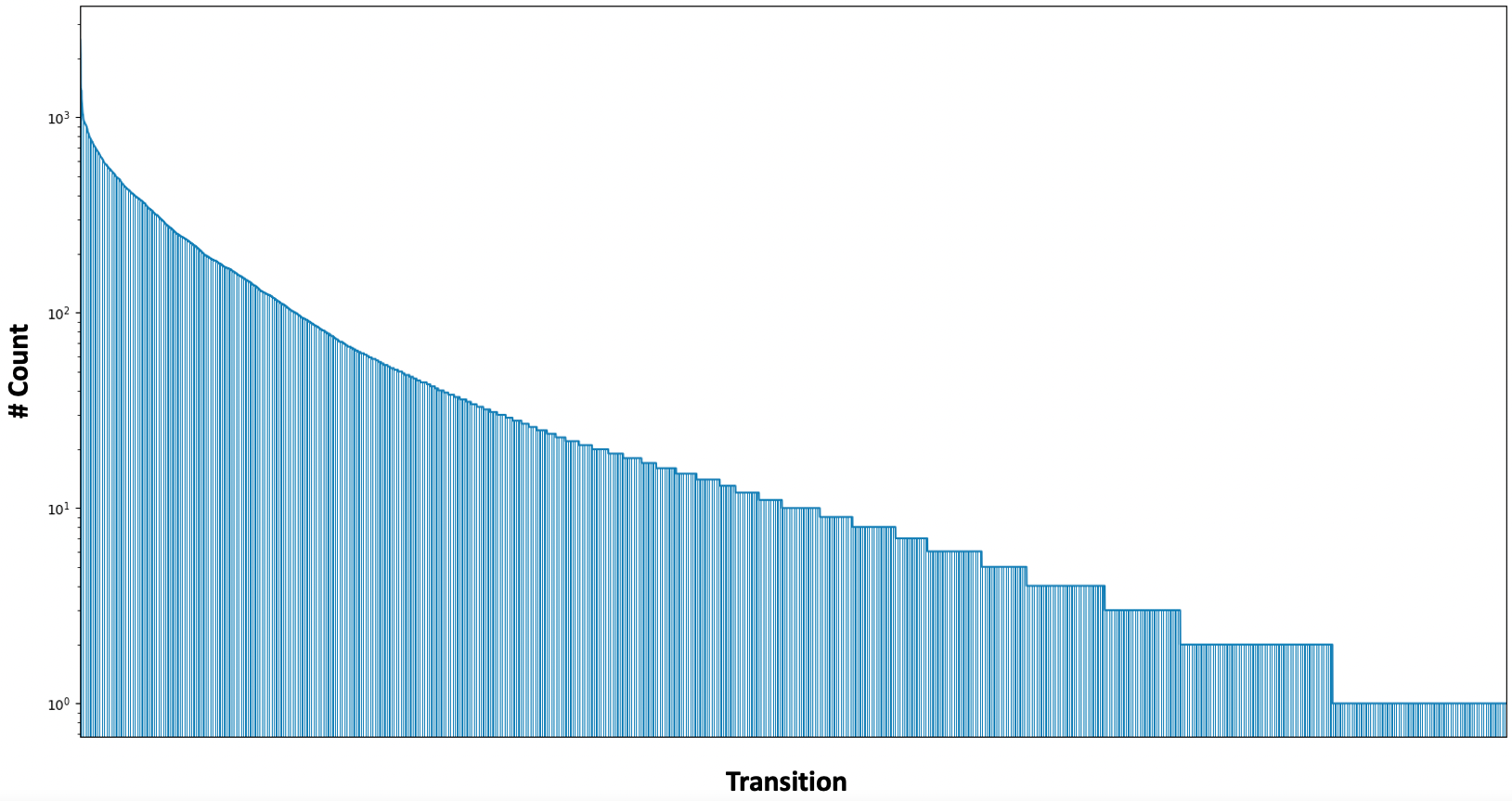}\\
    \caption{A visualization of all generated transitions (9,543). X axis is the transitions (e.g., \emph{jumping} changed to \emph{swimming}), and Y axis is logarithmic count.}
    \label{fig:transitions}
\end{figure}

\subsection{Additional Figures}
\label{sec:additional_figures}


\begin{figure}[!h]
\centering
\newcommand{\figlen}[0]{\columnwidth}
    \includegraphics[width=0.95\figlen]{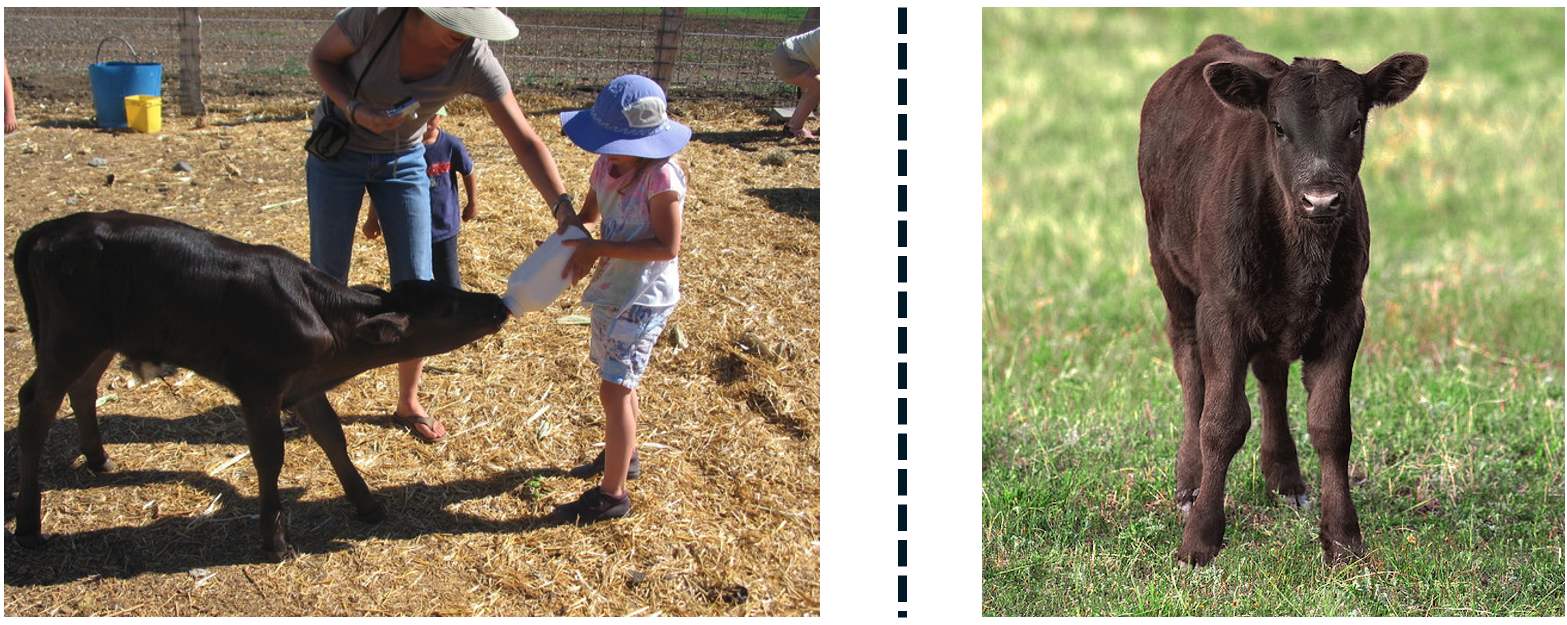}\\
    \caption{VASR focuses on complex images describing scenes, such as the image on the left (a child feeding a calf), rather than simpler images such as the image on the right.}
    \label{fig:single_object_vs_scene}
\end{figure}

\begin{figure}[!tb]
\centering
    \includegraphics[width=0.8\columnwidth]{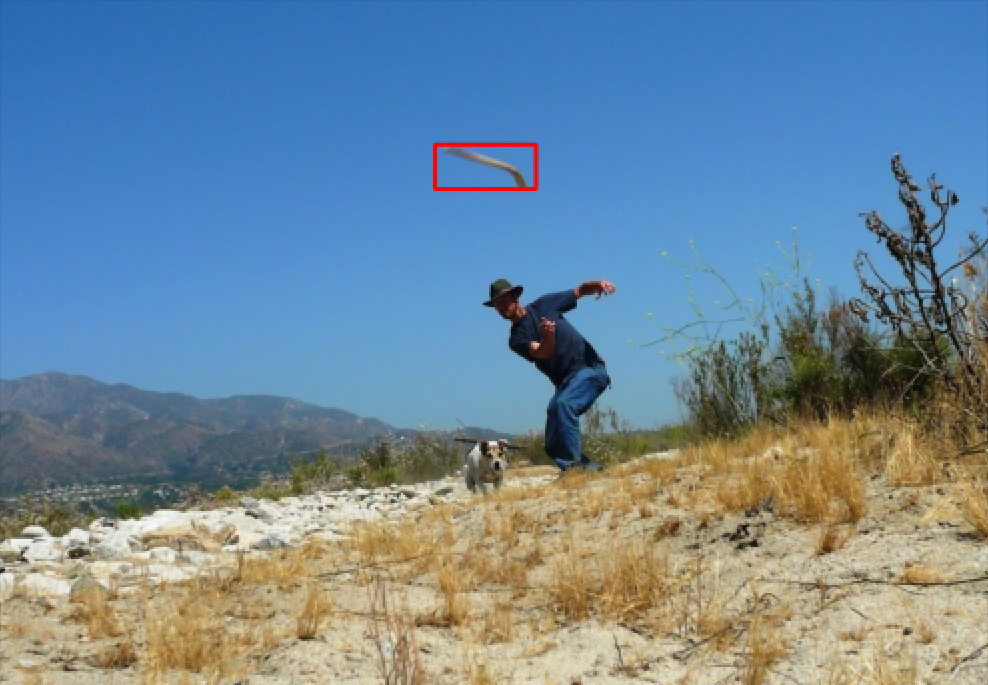}\\
    \caption{An example of non-visually salient object (2\% of the image), which we aim to filter from VASR.}
    \label{fig:fig_non_visually_salient}
\end{figure}

\begin{figure}[!h]
\centering
\newcommand{\figlen}[0]{\columnwidth}
    \includegraphics[width=0.95\figlen]{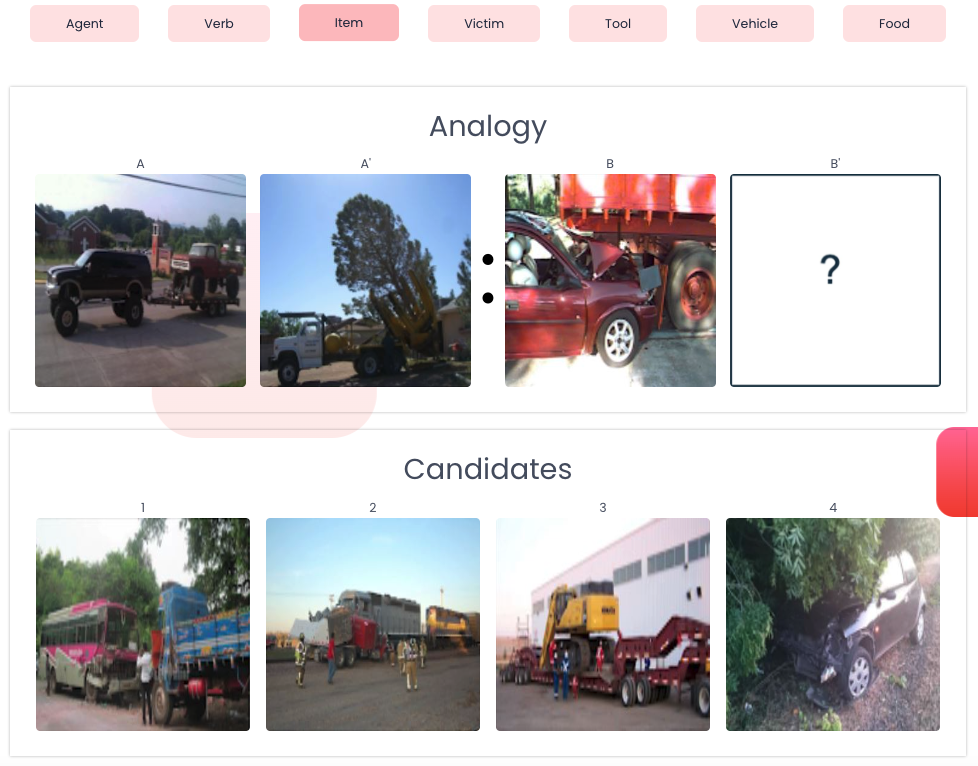}\\
    \caption{An example from VASR website that allows to users to interactively explore the different analogies in VASR. The following example presents an analogy of type \emph{item}.}
    \label{fig:explore_example}
\end{figure}

\begin{figure}[!h]
\centering
\newcommand{\figlen}[0]{\columnwidth}
    \includegraphics[width=0.95\figlen]{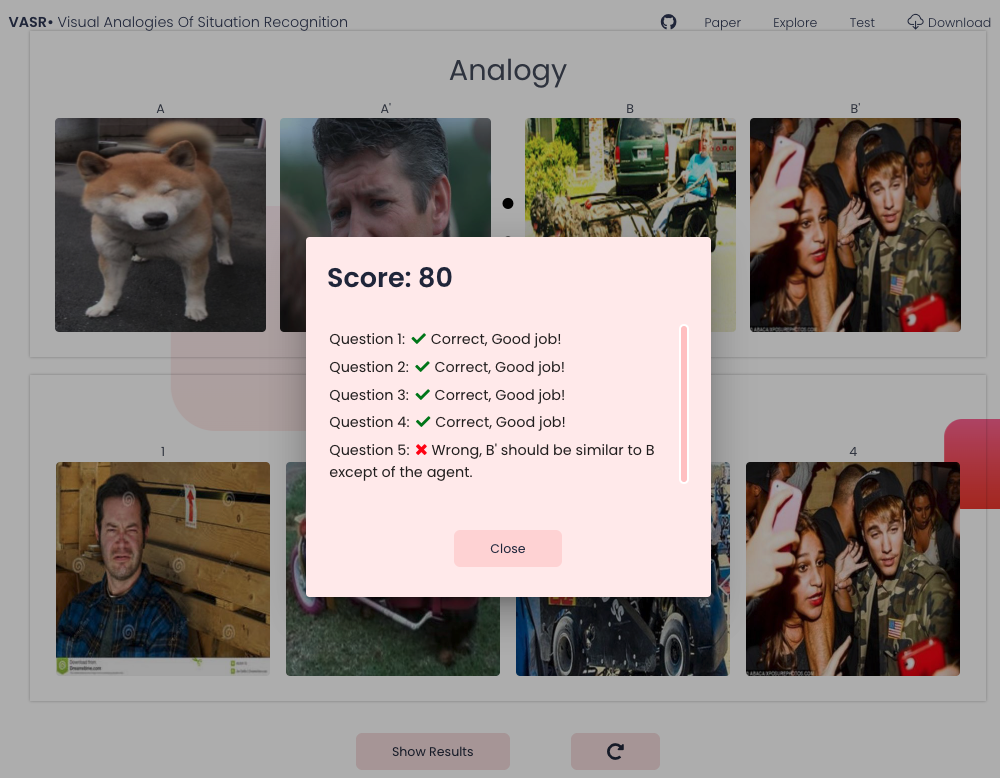}\\
    \caption{An example from VASR website that allows users to interactively solve analogies, receiving a grade and a feedback.}
    \label{fig:explore_test}
\end{figure}

\begin{figure}[!h]
\centering
\newcommand{\figlen}[0]{\columnwidth}
    \includegraphics[width=0.95\figlen]{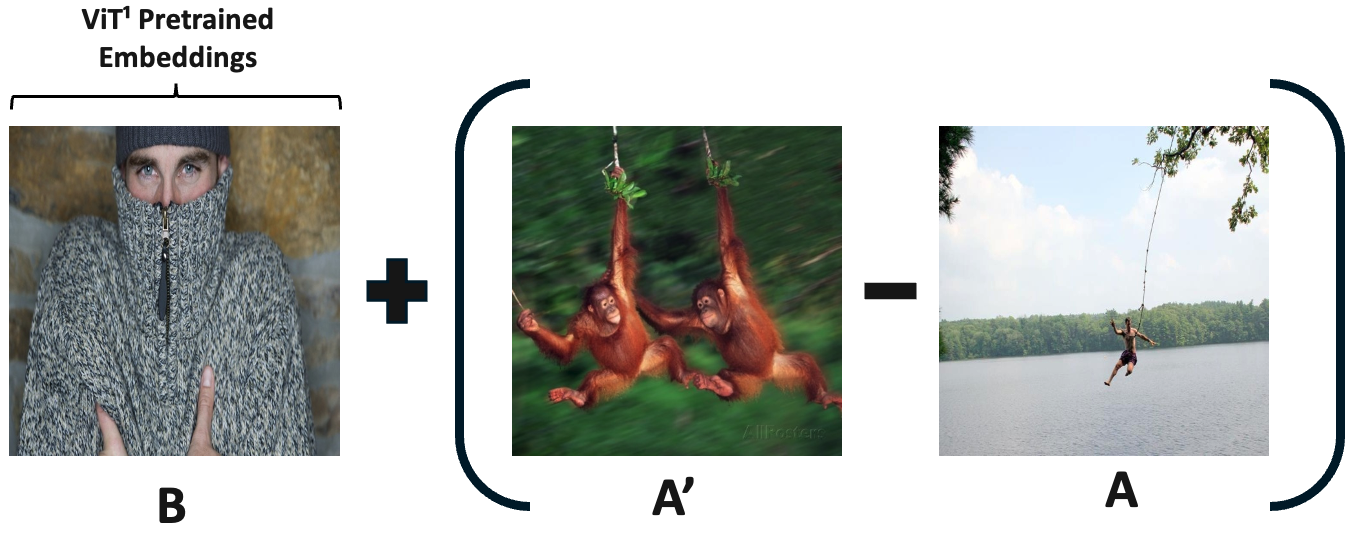}\\
    \caption{Zero-shot arithmetic model sketch. Given four candidates $C_1, C_2, C_3, C_4$, \newline$prediction = argmax_{i}(sim(B+(A'-A),C_i))$\newline The pretrained embeddings are obtained from some pretrained model, such as ViT, Swin Transformer, DEiT and ConvNeXt. We perform vector arithmetic $B + A' - A$, and select the candidate that is most similar (cosine-similarity) to the received representation.}
    \label{fig:zero_shot_arithmetics}
\end{figure}

\begin{figure}[!h]
\centering
\newcommand{\figlen}[0]{\columnwidth}
    \includegraphics[width=0.95\figlen]{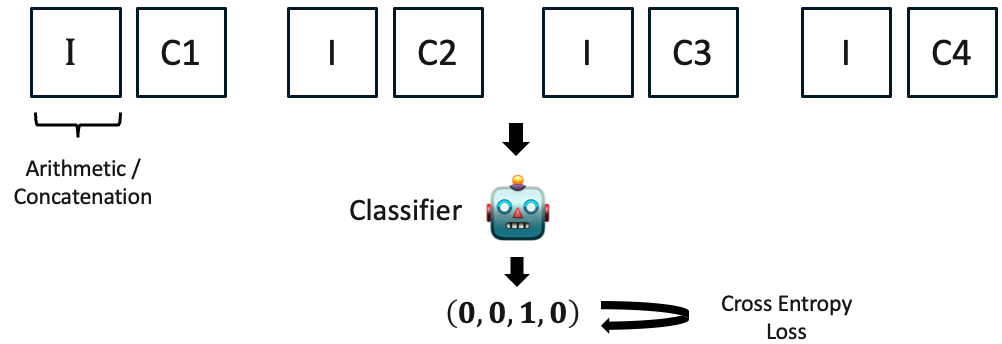}\\
    \caption{Supervised model sketch. We denote with ``I'' the input representation, which can be both the arithmetic representation ($B+A'-A$) or the concatenation representation ($A,A',B$). To classify an image out of four candidates, we concatenate the input to each of the candidates, receiving an output vector, and extracts the cross-entropy loss to train the model.}
    \label{fig:supervised_model}
\end{figure}

\subsection{WordNet Concepts}
We use the following list, covering most of the objects annotations in imSitu:

animal, person, group, male, female, creation, wheeled vehicle, system of measurement, structure, phenomenon, covering, celestial body, food, furniture, body of water, instrumentality, geographical area, round shape, plant, fire, tube, educator, liquid, leaf, figure, substance, volcanic eruption, natural elevation, force, bird of prey, bovine, skeleton, male, female, body part, conveyance, utensil, dog, cat, rock, hoop, way, spiritual leader, spring, doll, plant part, piece of cloth, piece of cloth, plant organ, edible fruit, cord,jewelry, baseball, poster, javelin, cement, fabric, snow, football, ice, tape, screen, grave, plate, plastic, egg, collar, ribbon, rope, wool, glass, lumber, cake, powder, sink, balloon, mushroom

\end{document}